\documentclass[a4paper,fleqn]{cas-dc}

\usepackage[numbers]{natbib}

\usepackage{microtype}
\usepackage{graphicx}
\usepackage{subfigure}
\usepackage{booktabs} 
\usepackage{subfloat}

\usepackage{amsmath}
\usepackage{amssymb}
\usepackage{mathtools}
\usepackage{amsthm}
\usepackage{bm}
\usepackage{multirow} 
\usepackage{algorithmic}
\usepackage{algorithm}
\usepackage[section]{placeins}     
\usepackage{stfloats}              
\usepackage[capitalize,noabbrev]{cleveref}
\usepackage[textsize=tiny]{todonotes}
\usepackage{cuted}   

\usepackage{hyperref}
\hypersetup{colorlinks=true, linkcolor=blue, anchorcolor=blue, citecolor=blue}
\graphicspath{{figs/}}

\def\tsc#1{\csdef{#1}{\textsc{\lowercase{#1}}\xspace}}
\tsc{WGM}
\tsc{QE}
\tsc{EP}
\tsc{PMS}
\tsc{BEC}
\tsc{DE}

\begin{document}

\let\WriteBookmarks\relax
\def\floatpagepagefraction{1}
\def\textpagefraction{.001}
\let\printorcid\relax        
\shortauthors{Z. Jibao et~al.}

\title [mode = title]{Inner-Instance Normalization for Time Series Forecasting}                      

\author[1]{Zipo Jibao} 
\credit{Conceptualization of this study, Methodology, Writing – original draft, Visualization, Formal analysis, Data curation}
\affiliation[1]{organization={School of Science, Harbin Institute of Technology},
                city={Shenzhen},
                postcode={518055}, 
                state={Guangdong},
                country={China}}
                
\author[1]{Yingyi Fu} 
\credit{Visualization, Conceptualization of this study, Methodology, Data curation Software}

\author[1]{Xinyang Chen} 
\credit{Conceptualization of this study, Methodology, Software, Validation}

\author[2]{Guoting Chen} 
\ead{guoting.chen@univ-lille.fr}
\credit{Writing - Original draft preparation, Supervision, Project
 administration, Funding acquisition}
\affiliation[2]{organization={School of Science, Great Bay University},
                city={Dongguan},
                postcode={523000}, 
                state={Guangdong},
                country={China}}
\cormark[1]
\cortext[cor1]{Corresponding author}

\begin{abstract}
Real-world time series are influenced by numerous factors and exhibit complex non-stationary characteristics. Non-stationarity can lead to distribution shifts, where the statistical properties of time series change over time, negatively impacting model performance. Several instance normalization techniques have been proposed to address distribution shifts in time series forecasting. However, existing methods fail to account for shifts within individual instances, leading to suboptimal performance. To tackle inner-instance distribution shifts, we propose two novel point-level methods: Learning Distribution (LD) and Learning Conditional Distribution (LCD). LD eliminates internal discrepancies by fitting the internal distribution of input and output with different parameters at different time steps, while LCD utilizes neural networks to predict scaling coefficients of the output. We evaluate the performance of the two methods with various backbone models across public benchmarks and demonstrate the effectiveness of the point-level paradigm through comparative experiments. The code and datasets are available at \href{https://anonymous.4open.science/r/Anonymous-Code-for-Paper}{https://anonymous.4open.science/r/Anonymous-Code-for-Paper}.
\end{abstract}


\maketitle

\section{Introduction}

Real-world time series often exhibit distribution shifts, rendering the independent and identically distributed assumption invalid and resulting in suboptimal performance in the forecasting task. Normalization methods separate the unique characteristics of individual instances and the shared features of the entire time series, allowing forecasting models to capture the essential time dependence of the entire time series more precisely. The most prevalent approach involves separating the mean and variance from the original input sequence and, after prediction, reintegrating the mean and variance through denormalization \cite{UlyanovVL16}. 

It is important to recognize that the statistics of the horizon, such as mean and variance, differ from those of the lookback, introducing a distribution shift between the input and the target that the backbone model must handle. To overcome this challenge, researchers have proposed methods such as Dish-TS \cite{Fan23} and SAN \cite{Liu23}, which explicitly predict the statistics of the target sequence. In particular, Liu et al. recognized the existence of shifts within instances and employed a slice approach to mitigate them. Fan \cite{Fan23} categorized distribution shifts into two types: \textbf{Intra-space shift}, which refers to distribution shifts across different input sequences, including training and testing inputs, and \textbf{Inter-space shift}, which denotes distribution shifts between input and target sequences.

\begin{figure}[ht]\rmfamily       
\begin{center}
\centerline{\includegraphics[width=\columnwidth]{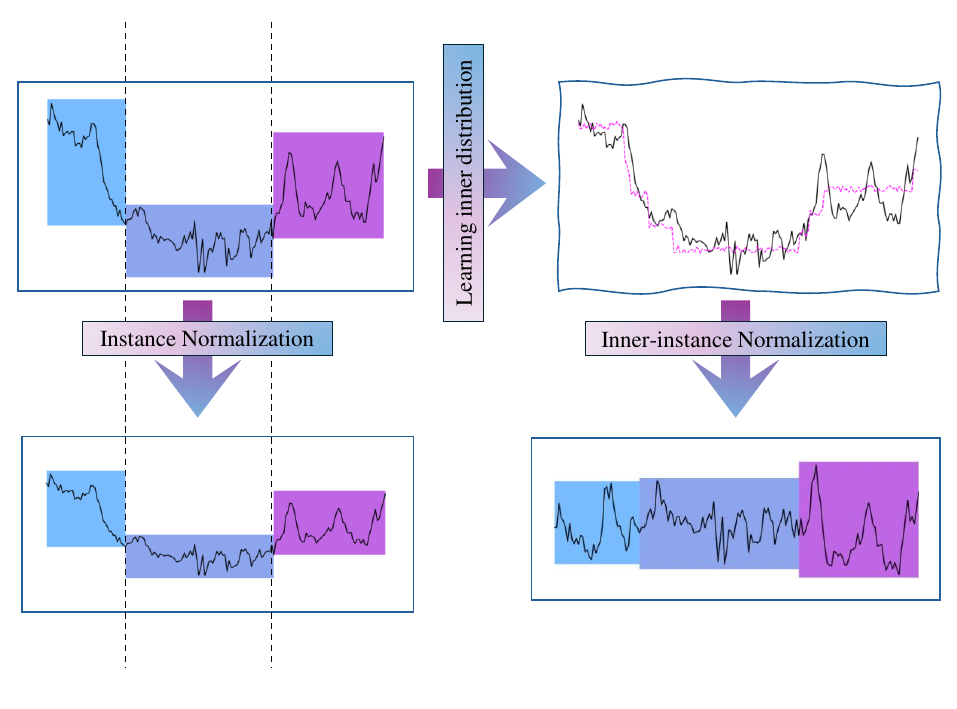}}
\caption{Point-level inner instance normalization could successfully alleviate inner-instance shifts. The example is from ETTh1.} 
\label{inner-shift}
\end{center}
\vskip -0.1in
\end{figure}

We have identified a third form of distribution shift within instances, which we term as \textbf{inner-instance shift}. It is widely recognized that the time series distribution changes from the input sequence to the target sequence. The length of the input (or target) in long-sequence prediction tasks can exceed the sum of the input and output lengths typically seen in short-sequence predictions. This observation leads us to conclude that distribution shifts occur at different points within the input (or target) sequence itself. Upon reviewing the results of the existing instance normalization models, we found that they have not addressed the distribution shifts within instances. As illustrated in \ref{inner-shift}, instance normalization is unable to address inner shifts, since translation and scaling will not change the shape of the original instance. Moreover, the lengths of stationary fragments are not a fixed number, so slice-level normalization is suboptimal.  

To address the inner-instance distribution shift, we propose two novel methods: Learning Distribution (LD) and Learning Conditional Distribution (LCD). LD introduces learnable parameters to fit the internal distribution at each time point for both the input space $\mathcal{X}$ and the target space $\mathcal{Y}$\footnote{Specifically, we learn the residual expectation at each time step.}. LD first applies the z-score to eliminate the inter-instance distribution shift, then removes the learned expectation of $\mathcal{X}$ at each time step to mitigate the internal distribution shifts of the input. The normalized sequence is put into the backbone model. After obtaining the output series from the backbone model, we introduce the expectations for each time step of $\mathcal{Y}$ and reintegrate the instance-specific statistics that were removed during z-score normalization. With this framework, the input and output of the backbone model are free from both inter-instance and inner-instance distribution shifts. We evaluate the performance of LD across three different models and compare it to RevIN, a model that fits statistics without accounting for inner-instance shifts. Our experiments show that LD consistently outperforms RevIN, demonstrating the effectiveness of the point-level paradigm in addressing the inner-instance shift.

To alleviate distribution shifts between $\bm{x}$ and $\bm{y}$ during forecasting, we need to learn the conditional distribution $P(\mathcal{Y}|\mathcal{X})$. As $P(\mathcal{Y}|\mathcal{X})$ is removed from the target sequence, the prediction task could be simplified to stationary prediction. LCD learns the conditional distribution $P(\mathcal{Y}|\mathcal{X})$ by learning $\bm{S(\bm{x})}$ and $\bm{\hat{\mu}_y(x)}$\footnote{On one hand, $\bm{S}$ and $\bm{\hat{\mu}_y}$ are functions of $\bm{x}$; on the other hand, they are variables that determine the distribution of $\bm{\hat{y}}$. This is why we refer to the prediction of $\bm{S}$ and $\bm{\hat{\mu}_y}$ as Learning Conditional Distribution.}. To mitigate the inner-instance shifts rooted in the horizon, we use the centered lookback to predict the scaling coefficients for the horizon at each time point. We multiply the output of the backbone model by the scaling coefficients. Finally, we add the predicted mean of the target sequence, derived from the lookback, to translate the result, yielding the final prediction. Compared with the instance-level normalization method Dish-TS and the slice-level method SAN, the time-point-level LCD demonstrates superior performance in experiments.

In summary, our contributions are as follows:
\begin{itemize}
\item We introduce the concept of inner-instance distribution shift, which has previously been overlooked. We utilize a time-point-level approach, which can serve as a new paradigm for future research. 
\item We propose two novel time-point-level models to address the inner-instance shift: one that resolves it through fitting internal distribution and the other through learning coefficient networks. 
\item Our models are simple, plug-and-play, and efficient, offering a significant improvement in the performance of forecasting models for real-world applications.
\end{itemize}

\section{Related Work}
\label{related work}  

\textbf{Time Series Forecasting.} ARIMA \cite{Whittle1963, BOXG1968} and its variants are among the longest-standing and most widely applied statistical models for time series forecasting. Holt introduced an exponential smoothing method that achieved top performance in its time \cite{HOLT2004}. Despite their strong theoretical foundations, traditional models face challenges due to the numerous assumptions they make, which often do not align with real-world data. With the advent of increased computational power, deep learning models have gained popularity. The Recurrent Neural Network (RNN) is a deep recurrent framework designed for sequence modeling, offering a variety of predictive models \cite{Wen2017, LAIG2018, SALINASD2020}. However, due to the incompatibility with parallel computation in the autoregressive framework, RNNs have gradually been supplanted by sequence-to-sequence mechanisms, including convolutional neural networks \cite{LIUM2022, Zhang2024} and Transformers. In recent years, attention-based models have been extensively proposed, such as Informer \cite{ZHOUH2021}, Autoformer \cite{WUH2021}, iTransformer \cite{liu24itransformer}, and AMSFormer \cite{Liu25}. Decomposition techniques \cite{OREHSKINBN2020, Yu24, Deng24}, frequency enhancement methods \cite{ZHOUT2022}, and patching strategies \cite{NIEY2022, ZHANGY2022, Zhang24, CHEN24} have become prevalent in time series forecasting. Some researchers have also integrated traditional methods with deep learning models \cite{WOOG2022}. Furthermore, studies have shown that even the simplest linear networks can achieve impressive results in time series prediction \cite{ZENGA2023, Toner24}, while another research \cite{Tan24} has argued that language models are ineffective in this long-standing domain.

\textbf{Normalization for Time Series Forecasting.} AdaRNN \cite{Du21} addresses the Temporal Covariate Shift problem by partitioning and matching the training data to mitigate the distribution shift. RevIN \cite{Kim22} introduces an instance-level normalization method that applies the same parameters to align both input and output distributions. Dish-TS \cite{Fan23} develops a dual coefficient network that predicts not only the mean and variance of the future sequence but also the hidden mean and variance of the input sequence. SAN \cite{Liu23} uses a slicing technique to segment the target sequence, predicting the mean and variance for each segment, and incorporates a two-stage training process to pre-train the normalization model. NST \cite{LIUY2022} leverages the properties of the softmax function to preserve information from the input prior to normalization. SOLID \cite{CHENM2023} fine-tunes their normalization model using training samples with a similar context when testing. Several forecasting models that aim to address distribution shifts \cite{LiuY23, Zhang24w} have been proposed. Three late studies apply Fourier transform and wavelet transform to address distribution shifts in frequency domains \cite{Ye24, Piao24, Dai24}. FOIL \cite{liu24} mitigates the distribution shift through invariant learning. SIN \cite{han24} identifies that locally invariant and globally variable statistics are well-suited for normalization and proposes an interpretable normalization model.

\section{Proposed Method}
\label{proposed method}

Multivariate time series forecasting predicts the subsequent sequence $\bm{y} \in R^{H\times D}$ using the input sequence $\bm{x} \in R^{L\times D}$. Model-agnostic normalization methods typically normalize the input before passing it through the backbone forecasting model, and then denormalize the output to obtain the final prediction. We use $g_{\theta}$ to represent the backbone model where $\theta$ denotes all the learnable parameters. In our paper, the terms lookback and input are used interchangeably (which implies that horizon and target are also equivalent). Notations in \ref{notations} are used throughout the paper.

\begin{table}[h]\rmfamily     
\caption{Mathematical notations.} 
\label{notations}
\begin{center}
\begin{small}
\begin{tabular}{ll}
\toprule
NOTATION &  DESCRIPTION \\
\midrule
$D$   & The dimension of time series\\
$L$ / $H$   &The lookback/horizon window size\\
$\mathcal{X}$ / $\mathcal{Y}$ &The input/target space \\
$\bm{x}^i$ / $\bm{y}^i$    & The $i$-th lookback/horizon  \\
$\Tilde{\bm{x}}^i$ / $\Tilde{\bm{y}}^i$   &The input/output of backbone model \\ 
$\bm{\mu}_x$, $\bm{\sigma}_x$   &The statistics of lookback \\
$\bm{\hat{\mu}}_y$, $\bm{\hat{\sigma}}_y$   &The  predicted statistics of horizon \\
\bottomrule
\end{tabular}
\end{small}
\end{center}
\vskip -0.1in
\end{table}

\subsection{Learning Distribution}
\label{LD}

LD learns the inner distributions of $\mathcal{X}$ and $\mathcal{Y}$ independently by fitting learnable matrices. We begin by applying z-score normalization to the input series. For the $i$-th sample $\bm{x}^i$, let $\bm{x}_{k}^{i}$ denote the $k$-th feature, and $x_{tk}^{i}$ the value at time step $t$. We first compute the mean and standard deviation of each feature as follows:
\begin{equation}\label{meanstd}
\mu_{k}^i = \frac{1}{L} \sum_{t=1}^{L} x_{tk}^{i},  \sigma_{k}^i = \sqrt{\frac{1}{L-1} \sum_{t=1}^{L}(x_{tk}^{i}-\mu_{k}^i)^2}.
\end{equation}
Next, we normalize each feature independently with the computed statistics:
\begin{equation}\label{z-score}
\overline{x}_{tk}^{i} = \frac{1}{\sigma_{k}^i + \epsilon} (x_{tk}^{i}-\mu_{k}^i),
\end{equation}
where $\epsilon$ is a small constant to avoid division by zero. At this point, the input series is preliminarily normalized. Under the assumption that the values of the input series over $L$ contiguous time steps are independent and identically distributed, the expectation of each $\overline{x}_{tk}^{i}$ is $0$ and its standard deviation is $1$, which implies successful normalization. However, in practice, time series exhibit significant drifts between adjacent time steps. Under this more realistic assumption, the expectation $\bm{A} \coloneqq E[\overline{\bm{x}}^i] \neq \bm{0}$ and the standard deviation matrix $\bm{B} \coloneqq \text{Std}[\overline{\bm{x}}^i] \neq \bm{J}$, where $\bm{J}$ is the $L \times D$ matrix with all entries equal to $1$. As a result, z-score normalization alone yields suboptimal results. To address this, we shift and scale $\overline{\bm{x}}^i$ using $\bm{A}$ and $\bm{B}$:
\begin{equation}\label{LDe}
\Tilde{\bm{x}}^i = (\overline{\bm{x}}^i - \bm{A}) \oslash \bm{B},
\end{equation}
where $\oslash$ means Hadamard division. To be more specific,

\begin{figure}[h] \rmfamily
    \centering
    \begin{footnotesize}
    \vspace*{1pt}
    \begin{equation}\label{LD-matrix}
    \bm{\Tilde{x}^i}
    =
    \begin{pmatrix}
    \frac{1}{b_{11}}(\overline{x}_{11}^{i}-a_{11}) & \frac{1}{b_{12}}(\overline{x}_{12}^{i}-a_{12}) & \cdots & \frac{1}{b_{1D}}(\overline{x}_{1D}^{i}-a_{1D}) \\
    \frac{1}{b_{21}}(\overline{x}_{21}^{i}-a_{21}) & \frac{1}{b_{22}}(\overline{x}_{22}^{i}-a_{22}) & \cdots & \frac{1}{b_{2D}}(\overline{x}_{2D}^{i}-a_{2D}) \\
    \vdots & \vdots & \ddots & \vdots \\
    \frac{1}{b_{L1}}(\overline{x}_{L1}^{i}-a_{L1}) & \frac{1}{b_{L2}}(\overline{x}_{L2}^{i}-a_{L2}) & \cdots & \frac{1}{b_{LD}}(\overline{x}_{LD}^{i}-a_{LD}) \\
    \end{pmatrix}.
    \end{equation}
    \end{footnotesize}
\end{figure}

According to the definition of $\bm{A}$, we have $E[\Tilde{\bm{x}}^i] = \bm{0}$, and $Std[\Tilde{\bm{x}}^i] = \bm{J}$, thus the distribution shifts within instance $\bm{x}^i$ is eliminated.

We have normalized the lookback sequence properly. Now we feed $\Tilde{\bm{x}}^i$ into the forecasting model as
\begin{equation}\label{forecast}
\Tilde{\bm{y}}^i = g_{\theta}(\Tilde{\bm{x}}^i), 
\end{equation}
and get $\Tilde{\bm{y}}^i$. Note there could be internal shifts in $\bm{y}^i$, we first denormalize $\Tilde{\bm{y}}^i$ using $\bm{P} \coloneqq E[\overline{\bm{y}}^i] \neq \bm{0}$ and $\bm{Q} \coloneqq Std[\overline{\bm{y}}^i]$:
\begin{equation}\label{LDeq}
\overline{\bm{y}}^i = \Tilde{\bm{y}}^i \odot \bm{Q} + \bm{P},
\end{equation}
where $\odot$ means Hadamard product. Finally, we use the statistics of the input $\bm{x}^i$ computed before to denormalize and get the final prediction:
\begin{equation}\label{LDequ}
\hat{y}_{nk}^{i} = \overline{y}_{nk}^{i} (\sigma_{k}^i + \epsilon) + \mu_{k}^i.
\end{equation}

If we disregard the means and standard deviations of the lookback, the entire procedure can be written into the following equation:
\begin{equation}\label{LDall}
(\mathcal{Y} - \bm{P}) \oslash \bm{Q} = g_{\theta}((\mathcal{X} - \bm{A}) \oslash \bm{B}).
\end{equation}
From \cref{LDall}, it is clear that two sets of matrices independently model the distributions of $\mathcal{Y}$ and $\mathcal{X}$. Because the normalization parameters vary across time steps, LD could individually adapt to the distribution at each time point, effectively addressing the issue of inner-instance distribution shift. Additionally, LD also explicitly mitigates inter-space shifts.

\begin{figure}[ht]\rmfamily      
\begin{center}
\centerline{\includegraphics[width=\columnwidth]{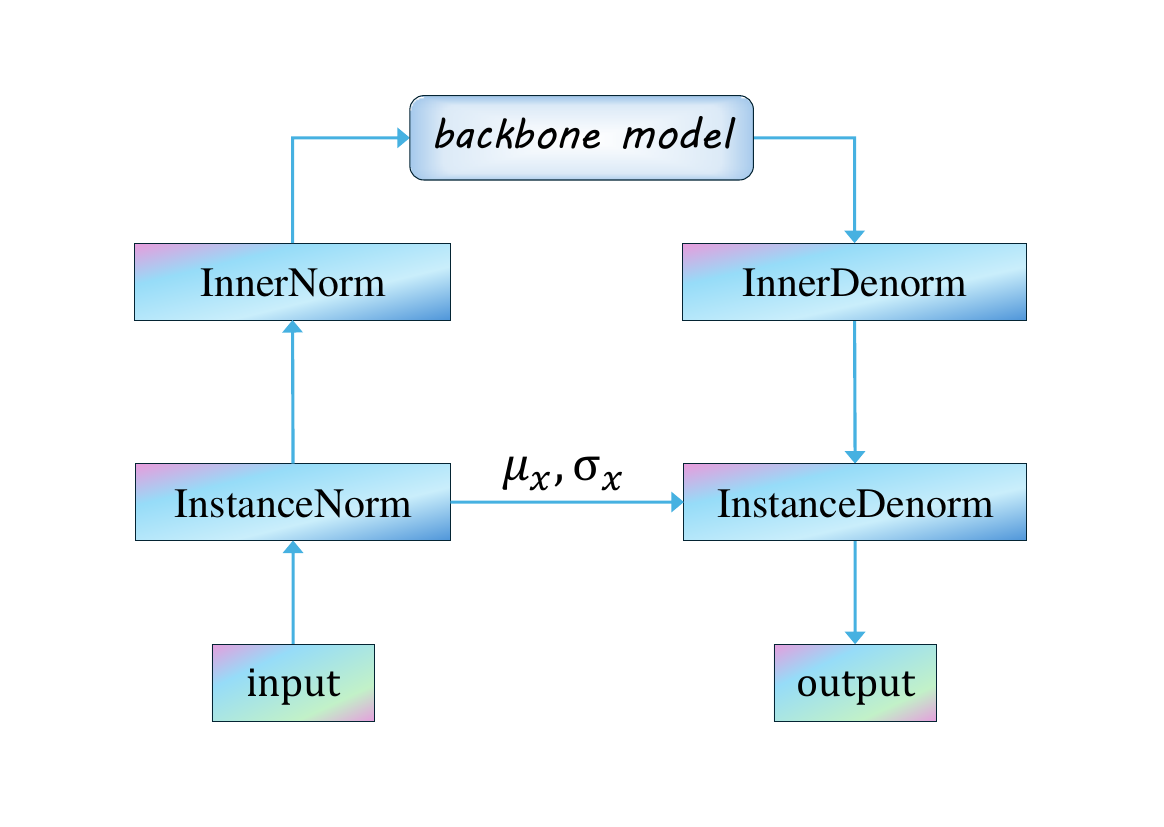}}
\caption{LD first eliminates the differences between instances and then resolves the inner distribution shifts within the instances.} 
\label{ld-frame}
\end{center}
\vskip -0.2in
\end{figure}

\cref{LDall} seems to indicate that the LD is complete to address normalization. Some readers may question the necessity of maintaining z-score normalization. As mentioned above, LD specifically addresses inner-instance shifts. Without z-score normalization, the distribution shifts between instances could hinder LD's ability to fit the internal distribution. The distribution shifts between instances encompass the entire original time series, while the internal instance distribution shifts only span $L+H$ time steps. Consequently, the distribution shifts between instances are the predominant factor. This is why LD normalization is implemented deeper inside the model than z-score normalization, illustrated in \ref{ld-frame}. 

Initially, we eliminate the instance-specific statistical information represented by $\bm{\mu^{i}_{k}}$ and $\bm{\sigma^{i}_{k}}$ from the lookback window. Subsequently, we remove the corresponding information for each time step to address the internal instance shifts, as $\bm{A}$ and $\bm{B}$ encapsulate time step information within the input and output. From the equation
\begin{equation}\label{z-score2}
\overline{x}_{tk}^{i} = \frac{1}{\sigma_{k}^i + \epsilon} (x_{tk}^{i}-\frac{1}{L} \sum_{t=1}^{L} x_{tk}^{i}),
\end{equation}
one could see that $\overline{x}_{tk}^{i}$ represents the residual between $x_{tk}^{i}$ and $\frac{1}{L} \sum_{t=1}^{L} x_{tk}^{i}$, effectively capturing the deviation of this time point from the instance mean. By fitting the mean and variance $x_{tk}$, both $\bm{A}$ and $\bm{B}$ effectively model the distribution of the differences between $x_{tk}$ and $\frac{1}{L} \sum_{t=1}^{L} x_{tk}$, thereby addressing the issue of inner-instance shift. Conversely, this theoretical framework is also applicable to $\bm{P}$ and $\bm{Q}$.In practical terms, we estimate the parameters by minimizing the loss function, as represented in the following equation:
\begin{equation}\label{LDtrain}
\mathop{\arg\min}\limits_{\theta, \bm{A}, \bm{P}} \sum_{(\bm{x^i}, \bm{y^i})} \mathcal{L}(\bm{y^i}, G(\bm{x^i}, g_{\theta}, \bm{A}, \bm{P})),
\end{equation}
where $G$ denotes the integrated forecasting model. During the gradient descent process aiming at minimizing the loss function, $\bm{A}$ and $\bm{B}$ gradually approximate the sample mean and standard deviation of $\bm{\overline{x}}$, thereby converging to the expectation and standard deviation of the random variable $\bm{\overline{x}}$. In practice, we find that it tends to overfit when applying both $\bm{A}$ and $\bm{B}$, so we remove the $\bm{B}$ part and set $\bm{A}$ as a matrix of learnable parameters. To some extent, the role of $\bm{B}$ can be played by weight matrices of neural networks, while $\bm{A}$ is irreplaceable. 

\subsection{Learning Conditional Distribution}
\label{LCD}

\begin{figure*}[b]\rmfamily
\vskip 0.1in
\begin{center}
\centerline{\includegraphics[width=0.9\linewidth]{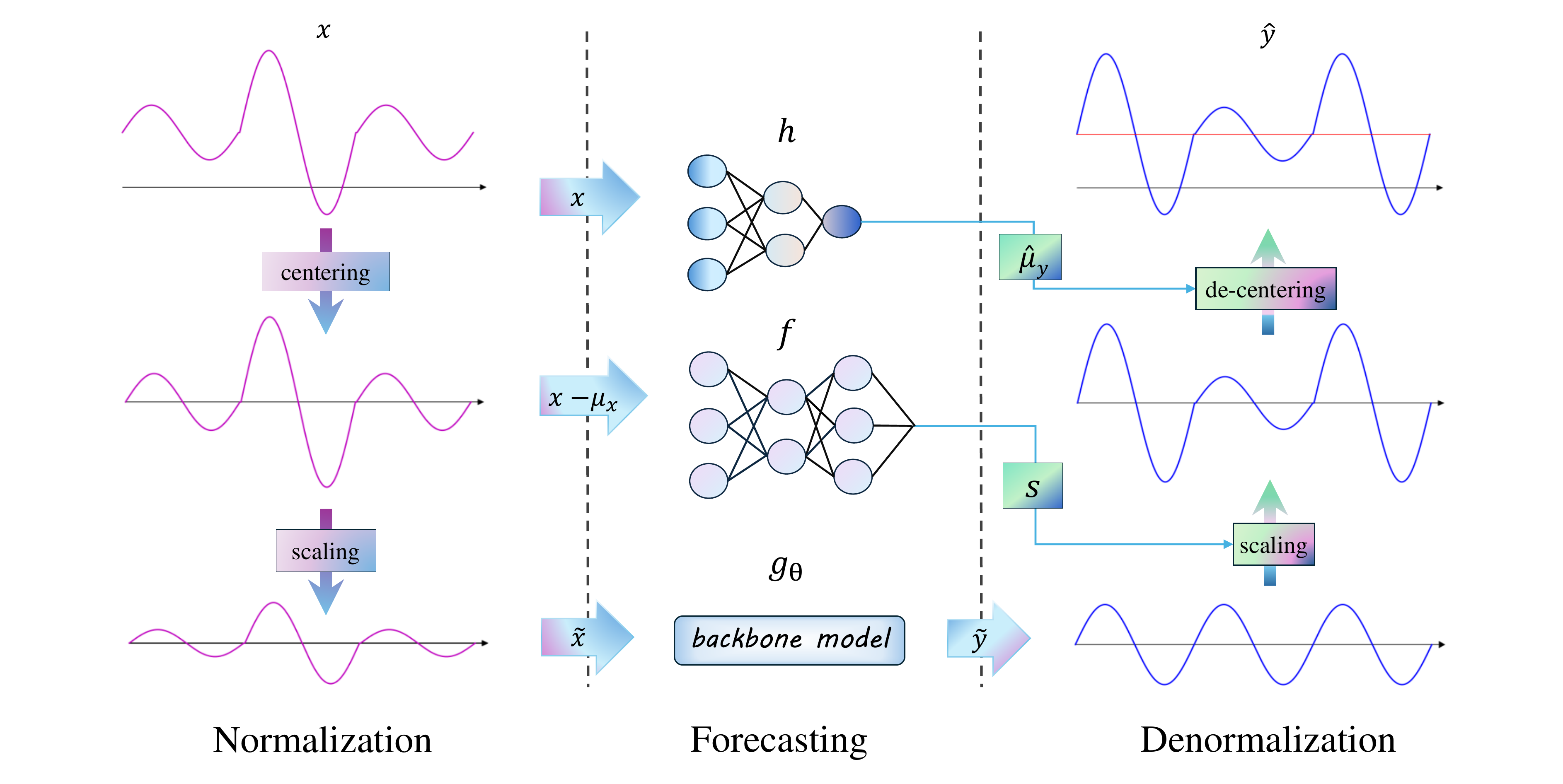}}
\caption{Overview of the proposed Learning Conditional Distribution framework, illustrating how LCD normalizes the non-stationary input, and how the information removed during normalization is processed and integrated into the output of the backbone model during de-normalization.}
\label{lcd-frame}
\end{center}
\vskip -0.2in
\end{figure*}

Time series forecasting involves learning a function $f(\mathcal{Y}|\mathcal{X})$ that relates random variables. Normalization methods aim to decouple the conditional distribution $P(\bm{y}|\bm{x})$ from the forecasting task, thereby minimizing the impact of distribution shifts on the backbone model. Existing methods typically predict the means and standard deviations of the horizon window. However, this approach does not adequately address internal distribution shifts within the horizon sequence. In contrast, the Learning Conditional Distribution (LCD) method is capable of predicting more fine-grained scaling coefficients and means. As illustrated in \ref{lcd-frame}, our central idea is that to eliminate inner-instance shifts, we only need to center the sequence and subsequently scale the centered values with different coefficients for each time point. To achieve this, we predict a future mean, denoted as $\bm{\hat{\mu}_y}$ to center the target, and a scale coefficient matrix $\bm{S}$ to perform the scaling. Then one can learn the conditional distribution through 

\begin{footnotesize}
\begin{equation}\label{lcd-decomp}
P(\bm{y}|\bm{x}) = P(\bm{y}|\bm{S}, \bm{\hat{\mu}_y}) P(\bm{S}, \bm{\hat{\mu}_y}|\bm{x}) = P(\bm{y}|\bm{S}, \bm{\hat{\mu}_y}) P(\bm{S}(\bm{x}), \bm{\hat{\mu}_y}(\bm{x})). 
\end{equation}
\end{footnotesize}

We assume that the time series is one-dimensional for simplicity, with an input sequence $\bm{x} = (x_1, x_2, \ldots, x_L)$ and a target sequence $\bm{y} = (y_1, y_2, \ldots, y_H)$. Having removed the mean from the input in the preprocessing stage, we must reintroduce it. To account for the distribution shift between lookbacks and horizons, we predict the future mean rather than utilizing the lookback mean directly. We denote this prediction as $h$, and the process is described by:
\begin{equation}\label{pre-muy}
\hat{\mu}_y = h(x_1, x_2, ..., x_L).
\end{equation}

It is reasonable to assert that the fluctuations in the target sequence are largely dependent on the fluctuations in the lookback sequence. Therefore, we utilize a centered lookback to predict future fluctuations. Specifically, we predict a scaling coefficient $s_n$ for the $n$-th future time step as 
\begin{equation}\label{pred-S}
s_n = f_n(x_1-\mu_x, x_2-\mu_x, ..., x_L-\mu_x), \quad n = 1, 2, ... , H. 
\end{equation}

We then scale and translate the output of the backbone model using:
\begin{equation}\label{pre-S}
\hat{y}_n = \Tilde{y}_n s_n + \hat{\mu}_y, \quad n = 1,2,...,H. 
\end{equation}

\begin{figure}[ht]\rmfamily      
\begin{center}
\centerline{\includegraphics[width=\columnwidth]{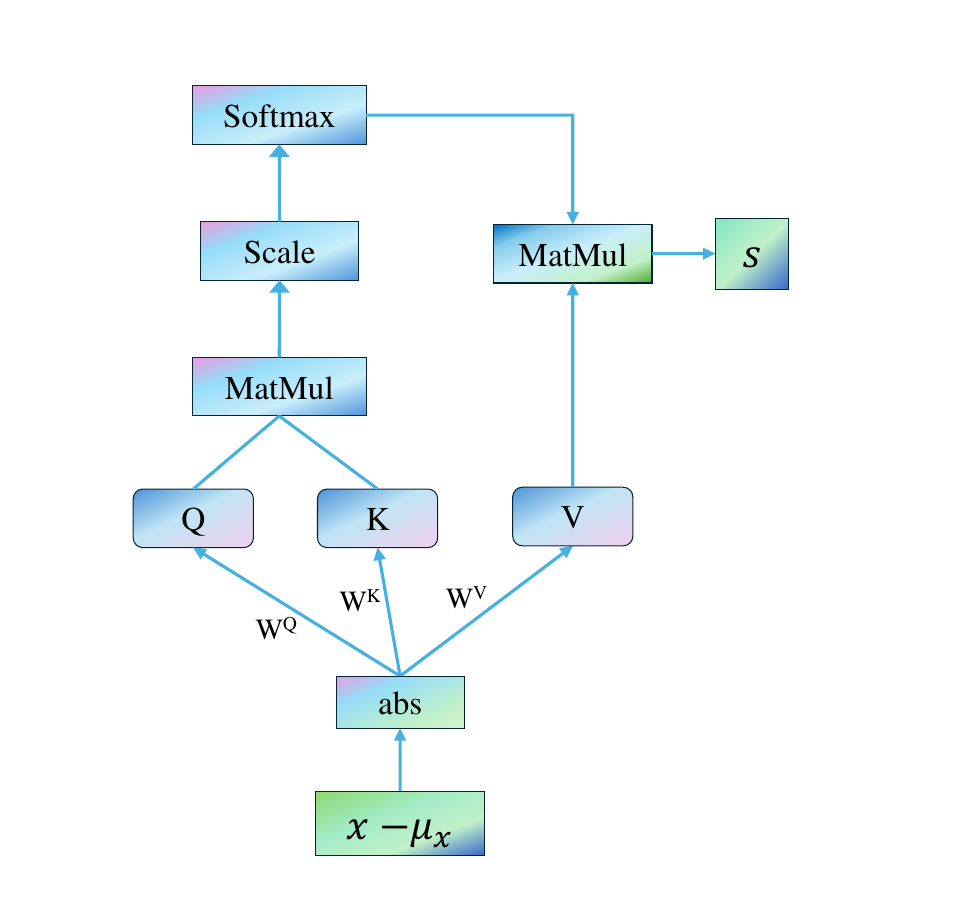}}
\caption{The network of LCD-as. We first calculate the absolute value of the centered input, and then calculate the attention score using the absolute value.} 
\label{lcd-as-frame}
\end{center}
\vskip -0.2in
\end{figure}

At this point, we have addressed the intra-space shift through z-score normalization, the inter-space shift through the matrix $\bm{S}$ and the predicted mean $\hat{\mu}_y$, as well as the inner-instance shift with $\bm{S}$. The functions $\bm{f} \coloneqq (f_1, f_2, ..., f_H)$ and $h$ can be implemented using any architecture, providing flexibility for the applications of our model. In this paper, we propose a linear network LCD-linear where $\bm{f}$ and $h$ are linear functions, and an attention score method LCD-as, which uses attention score as scaling coefficients $\bm{S}$, illustrated in \ref{lcd-as-frame}. 

Multivariate forecasting is a repetition of one-dimensional forecasting, using functions $h$ and $\bm{f}$ with shared parameters or different parameters for different features (i.e., dimensions). Let $\bm{\Tilde{y}}$ represent the output of the backbone model. The final prediction from LCD can be expressed as 
\begin{equation}\label{LCD-matrix}
\begin{footnotesize}
\bm{\hat{y}}
=
\begin{pmatrix}
\Tilde{y}_{11} s_{11} + \hat{\mu}_{y1} & \Tilde{y}_{12} s_{12} + \hat{\mu}_{y2} & \cdots & \Tilde{y}_{1D} s_{1D} + \hat{\mu}_{yD} \\
\Tilde{y}_{21} s_{21} + \hat{\mu}_{y1} & \Tilde{y}_{22} s_{22} + \hat{\mu}_{y2} & \cdots & \Tilde{y}_{2D} s_{2D} + \hat{\mu}_{yD} \\
\vdots & \vdots & \ddots & \vdots \\
\Tilde{y}_{L1} s_{L1} + \hat{\mu}_{y1} & \Tilde{y}_{L2} s_{L2} + \hat{\mu}_{y2} & \cdots & \Tilde{y}_{LD} s_{LD} + \hat{\mu}_{yD} \\
\end{pmatrix}.
\end{footnotesize}
\end{equation}
It is important to note that $\bm{S}$ and $\bm{\hat{\mu}_{y}}$ are functions of $\bm{X}$, which leads to a more precise relationship:
\begin{equation}\label{LCD-nk}
\hat{y}_{nk} =  \Tilde{y}_{nk}(\bm{x}) s_{nk}(\bm{x}) + \hat{\mu}_{yk}(\bm{x}).
\end{equation}

Because $s_{nk}(\bm{x})$ and $\hat{\mu}_{yk}(\bm{x})$ are predictable terms rather than statistics of the lookback window, LCD mitigates inter-space shifts. Assuming that the outputs from the backbone model $\Tilde{y}_{mk}(\bm{x})$ and $\Tilde{y}_{nk}(\bm{x})$ follow the same distribution, we can consider that $s_{mk}(\bm{x})$ and $s_{nk}(\bm{x})$ are functions distinct parameters, leading to the conclusion 
\begin{equation}\label{LCD-analysis}
    \begin{gathered}
        E[\hat{y}_{mk}] =  E[\Tilde{y}_{mk}(\bm{x}) s_{mk}(\bm{x}) + \hat{\mu}_{yk}(\bm{x})] \\
        \neq E[\Tilde{y}_{nk}(\bm{x}) s_{nk}(\bm{x}) + \hat{\mu}_{yk}(\bm{x})] = E[\hat{y}_{nk}], 
    \end{gathered}
\end{equation}
and
\begin{equation}\label{LCD-analysis2}
    \begin{gathered}
        Var[\hat{y}_{mk}] =  Var[\Tilde{y}_{mk}(\bm{x}) s_{mk}(\bm{x}) + \hat{\mu}_{yk}(\bm{x})] \\
        \neq Var[\Tilde{y}_{nk}(\bm{x}) s_{nk}(\bm{x}) + \hat{\mu}_{yk}(\bm{x})] = Var[\hat{y}_{nk}].
    \end{gathered}
\end{equation}

As a result, LCD enables the final predictions to adhere to different distributions, even when the outputs from the backbone model exhibit the same distribution at each time step. This effectively resolves the internal distribution discrepancy within instances.

It is also noteworthy that $s_{nk}(\bm{x})$ is a function of $\bm{x}$, indicating that 
$s_{nk}(\bm{x})$ and $\Tilde{y}_{nk}$ are not independent variables. Consequently, even if $s_{nk}(\bm{x})$ is a simple linear function of $\bm{x}$, the distribution of $\Tilde{y}_{nk}(\bm{x}) s_{nk}(\bm{x})$ remains complex. While $s_{nk}(\bm{x})$ captures information about both the mean and variance at time $t$, this information is inherently incomplete. To fully specify the mean and variance of the internal distribution over the horizon, at least $2HD$ variables would be necessary. However, we only utilize $D(H+1)$ variables in our framework. In this context of incompleteness, $s_{nk}(\bm{x})$ can flexibly adapt to learn the mean and variance based on specific circumstances.

\section{Experiments}
\label{experiments}

\subsection{Experimental Setup}
\label{setup}

\textbf{Backbone Models.} Our methods are designed as plug-and-play frameworks that are compact and effective. We utilize several state-of-the-art long-term forecasting models as backbones, including the linear model \textbf{DLinear}, CNN-based model \textbf{SCINet}, Transformer based model \textbf{Informer}, \textbf{Autoformer}, \textbf{FEDformer}, \textbf{iTransformer}, \textbf{PatchTST}, and \textbf{N-BEATS}, a hybrid model combining statistical methods with neural networks. During the experiment, we maintain all hyperparameters at their default values. Their code and be found at    
\begin{itemize}
\item \textbf{Informer}: \href{https://github.com/zhouhaoyi/Informer2020}{https://github.com/zhouhaoyi/Informer2020}.
\item \textbf{SCINet}: \href{https://github.com/cure-lab/SCINet}{https://github.com/cure-lab/SCINet}.
\item \textbf{N-BEATS}: \href{https://github.com/philipperemy/n-beats}{https://github.com/philipperemy/n-beats}.
\item \textbf{Autoformer}: \href{https://github.com/thuml/Autoformer}{https://github.com/thuml/Autoformer}.
\item \textbf{FEDformer}: \href{https://github.com/MAZiqing/FEDformer}{https://github.com/MAZiqing/FEDformer}.
\item \textbf{DLinear}: \href{https://github.com/cure-lab/LTSF-Linear}{https://github.com/cure-lab/LTSF-Linear}.
\item \textbf{PatchTST}: \href{https://github.com/yuqinie98/PatchTST}{https://github.com/yuqinie98/PatchTST}.
\item \textbf{iTransformer}: \href{https://github.com/thuml/iTransformer}{https://github.com/thuml/iTransformer}.
\end{itemize}

\textbf{Baseline Models.} We compare the two proposed methods against four state-of-the-art normalization models for time series forecasting. Our LD model is evaluated alongside \textbf{RevIN}, a distribution learning method, which normalizes instances by applying shifts and scales using two learnable vectors. LCD is compared with \textbf{Dish-TS} and \textbf{SAN}, both of which learn the conditional distribution $P(\mathcal{Y}|\mathcal{X})$. Notably, SAN addresses distribution discrepancies within instances and employs a slicing method to segment the input sequence and compute the means and standard deviations for each segment. Using these computed values, the mean and standard deviation for each target slice are predicted, effectively performing the denormalization. Dish-TS predicts the hidden statistics of both inputs and outputs simultaneously. Non-stationary Transformer (NST) injects information from the original time series into the output to prevent information loss caused by normalization. A summary of key information regarding the normalization models is presented in \ref{baselines} (where $P$ denotes the slice length).
\begin{itemize}
\item \textbf{RevIN}: \href{https://github.com/ts-kim/RevIN}{https://github.com/ts-kim/RevIN}.
\item \textbf{SAN}: \href{https://github.com/icantnamemyself/SAN}{https://github.com/icantnamemyself/SAN}.
\item \textbf{Dish-TS}: \href{https://github.com/weifantt/Dish-TS}{https://github.com/weifantt/Dish-TS}.
\item \textbf{NST}: \href{https://github.com/thuml/Nonstationary_Transformers}{Github: thuml/Nonstationary\_Transformers}.
\end{itemize}

\begin{table}[h]\rmfamily     
\caption{Information of normalization models}
\label{baselines}
\begin{center}
\begin{small}
\begin{tabular}{ccc}
\toprule
Model & Parameter & Granularity \\
\midrule
RevIN   & $2D$ & instance\\
Dish-TS   & $2DL$  & instance \\
SAN    & $1024(L+2H+PL)/P$  & slice\\
NST   & $6L+128(4D+256+1+L)$ & instance \\
LD     & $D(L+H)$ & time point \\
LCD-linear   & $DL(H+1)$  & time point \\ 
LCD-as   & $DL(3H+1)$  & time point \\ 
\bottomrule
\end{tabular}
\end{small}
\end{center}
\vskip -0.1in
\end{table}

\textbf{Datasets.} We evaluate the performance of the models using data sets from various domains, including energy (\textbf{ETT}), electricity (\textbf{Electricity}, or ECL), traffic (\textbf{Traffic}), finance (\textbf{Exchange}), and meteorology (\textbf{Weather}). Our multivariate time series benchmarks comprise real-world data, featuring a minimum of $7$ dimensions and a maximum of $862$ dimensions. The shortest series contains $966$ time steps, while the longest consists of $69680$ time steps. To prevent certain features from skewing the evaluation of experimental results and to improve readability, we standardize the entire dataset using the statistics derived from the training data. This setting does not result in the disclosure of subsequent test set information and aligns with the setup of RevIN and SAN. For all experiments, we partition the data into training, validation, and test sets in a $7:1:2$ ratio. The details are as follows.
\begin{itemize}
\item \textbf{Traffic:} contains hourly road occupancy rates from 862 sensors deployed on California highways in the United States, collected between 2015 and 2016. For more information, visit \href{https://pems.dot.ca.gov/}{https://pems.dot.ca.gov/}.
\item \textbf{Electricity:} records the electricity consumption of 321 clients, sampled every hour from 2012 to 2014. Additional details can be found at \\
\href{https://archive.ics.uci.edu/dataset/321/electricityloaddiagrams20112014}{https://archive.ics.uci.edu/dataset/321}.
\item \textbf{Weather:} retrieved from the Jena Beutenberg Weather Station in Germany includes 16 meteorological indicators, such as humidity and air pressure, with samples taken every ten minutes throughout the entire year of 2020. More information is available at \href{https://www.bgc-jena.mpg.de/wetter/}{https://www.bgc-jena.mpg.de/wetter/}.
\item \textbf{Exchange:} collects daily exchange rates for eight currencies, including the Australian dollar, the sterling, the Canadian dollar, the Swiss franc, the Chinese RMB, the yen, the New Zealand dollar, and the Singapore dollar, between 1990 and 2016. It is accessible at \\ \href{https://github.com/laiguokun/multivariate-time-series-data}{https://github.com/laiguokun/multivariate-time-series-data}.
\item \textbf{ETT-small:} comprises oil temperature readings of electrical transformers and six types of external power load features for two transformers at two stations in China, recorded from July 2016 to July 2018. The datasets ETTh1 and ETTh2 are sampled hourly. Further details can be found at \\
\href{https://github.com/zhouhaoyi/ETDataset}{https://github.com/zhouhaoyi/ETDataset}.
\end{itemize}

\begin{table*}[htb]\rmfamily
    \caption{Improvements on Informer, N-BEATS, and SCINet on 6 datasets, and comparison with state-of-the-art learning method RevIN. The last rows of the table represent the improvement rates of LD compared to vanilla backbone models. The best results are highlighted in bold. }
    \label{LDwo-tab3}
    \vskip 0.15in
    \setlength{\tabcolsep}{2pt}
    \begin{footnotesize}
    \centering
    \begin{tabular}{c|c|cccccc|cccccc|cccccc}
        \toprule
        \multirow{2}{*}{\rotatebox{90}{Data}}
        & Model & \multicolumn{2}{c}{Informer} & \multicolumn{2}{c}{+ LD} & \multicolumn{2}{c|}{+ RevIN} & \multicolumn{2}{c}{N-BEATS} & \multicolumn{2}{c}{+ LD} & \multicolumn{2}{c|}{+ RevIN} & \multicolumn{2}{c}{SCINet} & \multicolumn{2}{c}{+ LD}  & \multicolumn{2}{c}{+ RevIN} \\
        & Metric & MSE & MAE & MSE & MAE & MSE & MAE & MSE & MAE & MSE & MAE & MSE & MAE & MSE & MAE & MSE & MAE & MSE & MAE \\
        \midrule
        \multirow{6}{*}{\rotatebox{90}{Electricity}} 
        & 24   & 0.258 & 0.360 & \textbf{0.145} & \textbf{0.246} & 0.146 & 0.246 & 0.179 & 0.262 & \textbf{0.170} & \textbf{0.249} & 0.175 & 0.250 & 0.159 & 0.278 & \textbf{0.128} & \textbf{0.227} & 0.129 & 0.228  \\
        & 48   & 0.359 & 0.426 & \textbf{0.170} & \textbf{0.271} & 0.186 & 0.287 & 0.197 & 0.281 & \textbf{0.178} & \textbf{0.264} & 0.183 & 0.266 & 0.230 & 0.320 & \textbf{0.143} & \textbf{0.244} & 0.144 & 0.244 \\
        & 168  & 0.349 & 0.427 & \textbf{0.196} & \textbf{0.299} & 0.225 & 0.329 & 0.187 & 0.282 & \textbf{0.178} & \textbf{0.272} & 0.179 & 0.273 & 0.236 & 0.338 & \textbf{0.162} & \textbf{0.262} & 0.163 & 0.263 \\
        & 336  & 0.448 & 0.499 & \textbf{0.212} & \textbf{0.317} & 0.319 & 0.404 & 0.190 & 0.291 & \textbf{0.188} & \textbf{0.284} & 0.189 & 0.285 & 0.209 & 0.323 & \textbf{0.171} & \textbf{0.274} & 0.172 & 0.274 \\
        & 720  & 0.463 & 0.498 & \textbf{0.243} & \textbf{0.339} & 0.621 & 0.598 & \textbf{0.224} & 0.321 & 0.225 & 0.318 & 0.224 & \textbf{0.315} & 0.232 & 0.339 & 0.199 & 0.291 & \textbf{0.198} & \textbf{0.290} \\
        & Imp.  &  &  & 48.05\% & 31.12\% &  &  &  &  & 4.02\% & 3.58\% &  &  &  &  & 24.22\%  & 18.72\% &  &  \\
        \midrule
        \multirow{6}{*}{\rotatebox{90}{Traffic}} 
        & 24   & 0.607 & 0.334 & \textbf{0.588} & \textbf{0.314} & 0.627 & 0.341 & 0.614 & 0.365 & \textbf{0.553} & \textbf{0.333} & 0.567 & 0.337 & 0.524 & 0.322 & \textbf{0.480} & \textbf{0.298} & 0.484 & 0.301 \\
        & 48   & 0.677 & 0.379 & \textbf{0.622} & \textbf{0.341} & 0.926 & 0.476 & 0.630 & 0.370 & \textbf{0.589} & \textbf{0.360} & 0.594 & 0.362 & 0.568 & 0.355 & \textbf{0.502} & 0.320 & 0.503 & \textbf{0.319} \\
        & 168  & 0.728 & 0.401 & \textbf{0.636} & \textbf{0.345} & 1.043 & 0.547 & 0.563 & 0.354 & \textbf{0.510} & \textbf{0.337} & 0.514 & 0.339 & 0.554 & 0.358 & \textbf{0.486} & \textbf{0.332} & 0.487 & 0.334 \\
        & 336  & 0.867 & 0.477 & \textbf{0.662} & \textbf{0.366} & 1.048 & 0.558 & 0.531 & 0.342 & \textbf{0.476} & \textbf{0.326} & 0.478 & 0.326 & 0.520 & 0.339 & \textbf{0.475} & \textbf{0.325} & 0.476 & 0.326 \\
        & 720  & 1.026 & 0.556 & \textbf{0.711} & \textbf{0.396} & 1.113 & 0.574 & 0.544 & \textbf{0.346} & 0.499 & 0.349 & \textbf{0.495} & 0.349 & 0.527 & 0.366 & 0.498 & 0.350 & \textbf{0.494} & \textbf{0.347} \\
        & Imp.  &  &  & 15.55\% & 16.41\% &  &  &  &  & 8.89\% & 4.02\% &  &  &  &  & 9.25\%  & 6.53\% &  &  \\
        \midrule
        \multirow{6}{*}{\rotatebox{90}{ETTh1}} 
        & 24   & 0.575 & 0.525 & \textbf{0.516} & \textbf{0.483} & 0.547 & 0.485 & 0.400 & 0.419 & \textbf{0.384} & \textbf{0.404} & 0.387 & 0.406 & \textbf{0.391} & 0.423 & 0.406 & \textbf{0.421} & 0.419 & 0.426 \\
        & 48   & \textbf{0.574} & 0.538 & 0.663 & \textbf{0.535} & 0.658 & 0.536 & 0.407 & 0.430 & \textbf{0.405} & \textbf{0.424} & 0.406 & 0.425 & 0.417 & 0.434 & \textbf{0.410} & \textbf{0.421} & 0.412 & 0.423 \\
        & 168  & 0.814 & 0.661 & \textbf{0.607} & \textbf{0.564} & 0.671 & 0.570 & \textbf{0.489} & 0.499 & 0.498 & \textbf{0.498} & 0.499 & 0.499 & 0.542 & 0.524 & \textbf{0.493} & \textbf{0.485} & 0.494 & 0.486 \\
        & 336  & 1.386 & 0.861  & 1.336 & 0.860 & \textbf{0.749} & \textbf{0.616} & \textbf{0.556} & \textbf{0.551} & 0.586 & 0.558 & 0.588 & 0.559 & 0.661 & 0.605 & \textbf{0.530} & \textbf{0.517} & 0.532 & 0.520 \\
        & 720  & 1.465 & 0.897 & \textbf{1.134} & \textbf{0.791} & 1.207 & 0.827 & \textbf{0.719} & \textbf{0.643} & 0.768 & 0.670 & 0.765 & 0.667 & 1.104 & 0.794 & 0.709 & 0.642 & \textbf{0.708} & \textbf{0.640} \\
        & Imp.  &  &  & 9.28\% & 10.74\% &  &  &  &  & -1.91\% & 0.03\% &  &  &  &  & 12.42\%  & 8.86\% &  &  \\
        \midrule
        \multirow{6}{*}{\rotatebox{90}{ETTh2}} 
        & 24   & 0.169 & 0.296  & 0.142 & 0.260 & \textbf{0.137} & \textbf{0.252} & 0.124 & 0.241 & \textbf{0.123} & \textbf{0.236} & 0.124 & 0.237 & 0.127 & 0.250 & \textbf{0.119}  & \textbf{0.233} & 0.120 & 0.234 \\
        & 48   & 0.283 & 0.375 & \textbf{0.180} & \textbf{0.288} & 0.203 & 0.315 & 0.154 & 0.269 & \textbf{0.149} & \textbf{0.261} & 0.150 & 0.261 & 0.158 & 0.281 & \textbf{0.146} & 0.260 & 0.146 & \textbf{0.259} \\
        & 168  & 0.464 & 0.479 & \textbf{0.276} & \textbf{0.378} & 0.281 & 0.380 & 0.212 & 0.324 & 0.202 & \textbf{0.313} & \textbf{0.200} & 0.314 & 0.228 & 0.344 & \textbf{0.207}  & \textbf{0.313} & 0.208 & 0.314 \\
        & 336  & 0.554 & 0.567  & \textbf{0.246} & \textbf{0.355} & 0.289 & 0.388 & \textbf{0.217} & 0.336 & 0.234 & 0.340 & 0.220 & \textbf{0.335} & 0.248 & 0.375 & \textbf{0.230}  & \textbf{0.337} & 0.231 & 0.338 \\
        & 720  & 0.505 & 0.543 & \textbf{0.341} & \textbf{0.410} & 0.489 & 0.476 & \textbf{0.254} & \textbf{0.367} & 0.309 & 0.403 & 0.312 & 0.405 & 0.356 & 0.444 & 0.274 & 0.376 & \textbf{0.270}  & \textbf{0.373} \\
        & Imp.  &  &  & 36.27\% & 23.66\% &  &  &  &  & -4.14\% & -0.51\% &  &  &  &  & 10.68\%  & 9.82\% &  &  \\
        \midrule
        \multirow{6}{*}{\rotatebox{90}{Weather}} 
        & 24   & 0.101 & 0.203 & \textbf{0.078} & \textbf{0.150} & 0.080 & 0.157 & 0.063 & 0.127 & \textbf{0.061} & \textbf{0.118} & 0.062 & 0.119 & 0.073 & 0.154 & \textbf{0.060}  & \textbf{0.118} & 0.061 & 0.120 \\
        & 48   & 0.291 & 0.371 & \textbf{0.133} & \textbf{0.216} & 0.151 & 0.233 & 0.109 & 0.193 & \textbf{0.109} & \textbf{0.185} & 0.110 & 0.187 & 0.161 & 0.254 & \textbf{0.118}  & \textbf{0.194} & 0.119 & 0.195 \\
        & 168  & 0.248 & 0.350 & \textbf{0.219} & \textbf{0.301} & 0.222 & 0.303 & 0.181 & 0.277 & \textbf{0.177} & \textbf{0.265} & 0.179 & 0.266 & 0.202 & 0.306 & \textbf{0.175} & \textbf{0.266} & 0.178  & 0.269 \\
        & 336  & 0.404 & 0.445 & \textbf{0.335} & \textbf{0.376} & 0.342 & 0.380 & 0.269 & 0.349 & \textbf{0.255} & \textbf{0.327} & 0.256 & 0.329 & 0.288 & 0.371 & \textbf{0.255}  & \textbf{0.328} & 0.258 & 0.329 \\
        & 720  & 0.517 & 0.512 & 0.490 & 0.463 & \textbf{0.413} & \textbf{0.427} & 0.418 & 0.438 & \textbf{0.394} & \textbf{0.416} & 0.403 & 0.420 & 0.436 & 0.467 & \textbf{0.365} & \textbf{0.402} & 0.374  & 0.407 \\
        & Imp.  &  &  & 22.20\% & 21.29\% &  &  &  &  & 3.26\% & 5.38\% &  &  &  &  & 16.64\%  & 16.34\% &  &  \\
        \midrule
        \multirow{6}{*}{\rotatebox{90}{Exchange}} 
        & 24   & 0.468 & 0.557 & \textbf{0.031} & \textbf{0.120} & 0.035 & 0.125 & 0.028 & 0.114  & \textbf{0.024} & \textbf{0.102} & 0.025 & 0.103 & 0.077 & 0.210 & \textbf{0.024}  & \textbf{0.104} & 0.024 & 0.104 \\
        & 48   & 0.790 & 0.735 & 0.064 & \textbf{0.178} & \textbf{0.061} & 0.179 & 0.054 & 0.166 & \textbf{0.041} & \textbf{0.140} & 0.043 & 0.142 & 0.151 & 0.301 & \textbf{0.040}  & \textbf{0.142} & 0.043 & 0.145 \\
        & 168  & 1.046 & 0.851 & \textbf{0.298} & \textbf{0.359} & 0.334 & 0.382 & 0.297 & 0.403 & 0.154 & 0.283 & \textbf{0.151} & \textbf{0.280} & 0.385 & 0.496 & \textbf{0.150}  & \textbf{0.283} & 0.151 & 0.284 \\
        & 336  & 1.454 & 1.023 & \textbf{0.378} & \textbf{0.450} & 0.693 & 0.599 & 0.637 & 0.618 & 0.368 & 0.450 & \textbf{0.360} & \textbf{0.446} & 1.031 & 0.822 & \textbf{0.352}  & \textbf{0.436} & 0.354 & 0.437 \\
        & Imp.  &  &  & 82.66\% & 63.11\% &  &  &  &  & 32.18\% & 20.79\% &  &  &  &  & 66.70\%  & 47.97\% &  &  \\
        \bottomrule
    \end{tabular}
    \end{footnotesize}
\vskip -0.1in
\end{table*}

\textbf{Implementation.} As aforementioned, we retain all hyperparameters of the backbone models at their respective default values. The batch size is consistently set to $128$ unless it is constrained by memory limitations, in which case it is adjusted to $64$. An exception occurs when using Autoformer to predict a sequence of length $720$, where the batch size is further reduced to $32$ due to GPU memory constraints.  The learning rate is set to $10^{-3}$ for transformer-based models and $10^{-4}$ for models from other families. However, for the Electricity and Traffic datasets, the learning rate is fixed at $10^{-4}$ to prevent gradient explosion, which can occur when set to $10^{-3}$. Some prior studies fixed the input sequence length to $96$ for all horizon lengths, while others used a larger input length for longer target series. Our findings indicate that the prediction error is minimized on average when the input length $L$ is set to \{$24, 48, 72, 96, 120, 192, 360$\} for the corresponding target length \{$24, 48, 96, 168, 192, 336, 720$\} for most models. Therefore, we maintain this correspondence for all experiments conducted. For the Transformer models, the decoder input length is consistently set to half the lookback window. When using transformer-based models as the backbone, the input for SAN includes temporal information. Our experiments demonstrated that the influence of temporal information on the results is negligible; consequently, when utilizing other normalization models, the input excludes temporal data. We employ the Adam optimizer and conduct all experiments on a computer equipped with an Intel i5-12600KF processor and an NVIDIA RTX 4060Ti 16GB GPU.

\subsection{LD Performance}\label{LDp}

\textbf{Improvements on models.} We analyze the performance variations before and after implementing LD across three backbone models: Informer, N-BEATS, and SCINet. We chose the three backbone models for RevIN used them as backbone models\cite{Kim22}. The results, illustrated in \ref{LDwo-tab3}, indicate that LD enhances the performance of all models across all datasets. Specifically, LD produces the most significant improvement for Informer, with average MSE reductions of $48.05\%$, $15.55\%$, $9.28\%$, $36.27\%$, $22.20\%$, and $82.66\%$ \footnote{The percentage is calculated for each prediction length and then averaged.} across the six datasets. In the best-case scenario, the prediction error for Informer in the Exchange dataset decreases by an average of $82.67\%$ following the application of LD. For N-BEATS, LD yields MSE error reductions of $4.02\%$, $8.89\%$, $-1.91\%$, $-4.14\%$, $3.26\%$, and $32.18\%$. Similarly, LD decreases prediction errors for the SCINet model by $24.22\%$, $9.25\%$, $12.42\%$, $10.68\%$, $16.64\%$, and  $66.70\%$. It is evident that as the prediction length increases and model prediction errors escalate, LD enhances the perfirmance stability. Notably, LD consistently improves performance on Exchange for all three models, demonstrating its effectiveness in mitigating distribution shifts, particularly given that Exchange exhibits the highest non-stationarity with an ADF score of $-1.9$.

\begin{figure*}[h]\rmfamily
    \centering
    \rotatebox{90}{\scriptsize{~~~~~~~~~~~~~~~~~~~~~~~\normalsize{ETTh1}}}
    \subfigure{
        \begin{minipage}[t]{0.44\linewidth}
            \centering
            \includegraphics[width=1\linewidth]{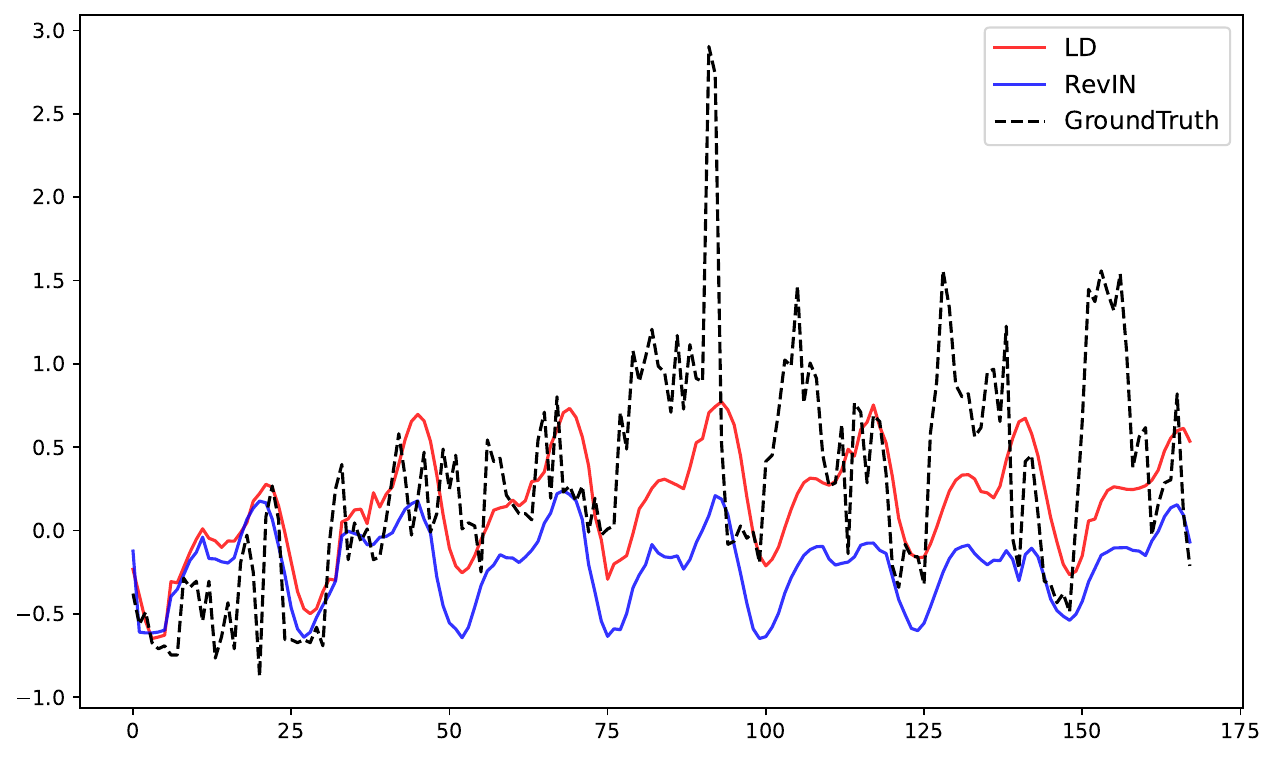}
        \end{minipage}
    }
    \subfigure{
        \begin{minipage}[t]{0.44\linewidth}
            \centering
            \includegraphics[width=1\linewidth]{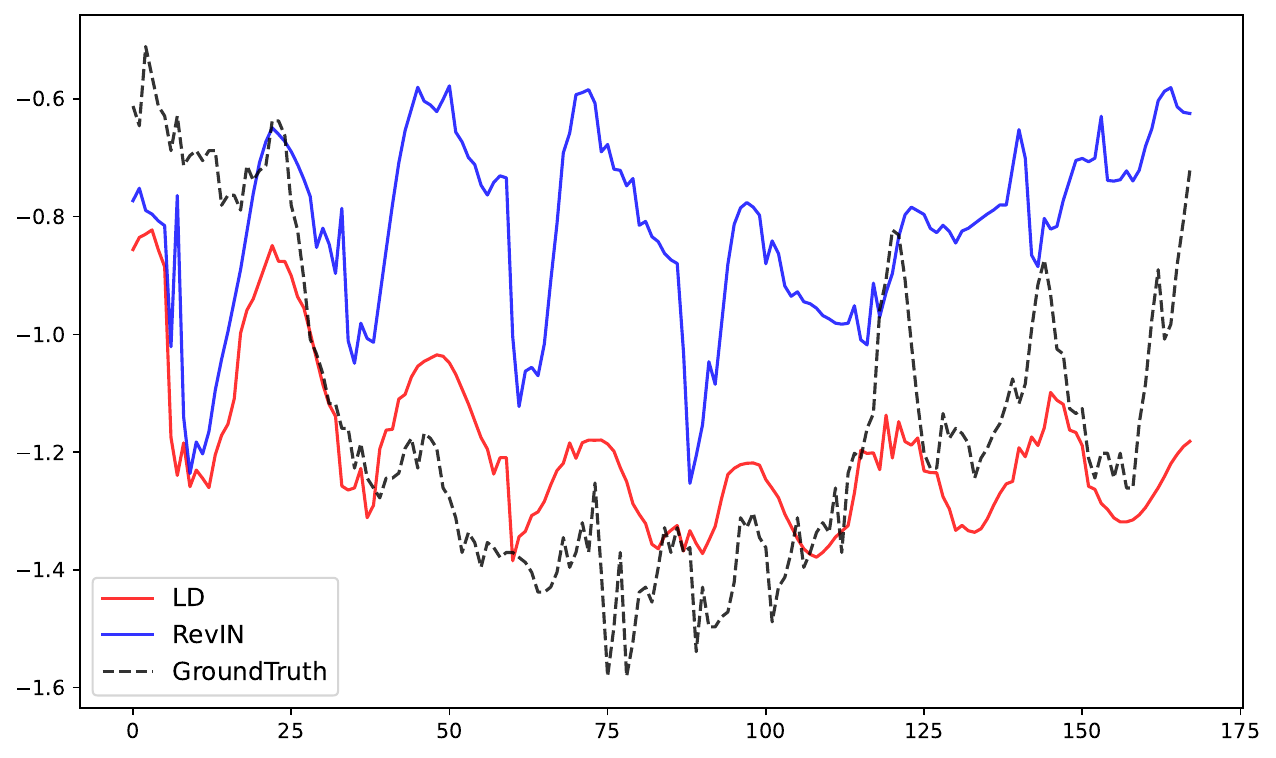}
        \end{minipage}
    }
    
    \setcounter{subfigure}{0}
     \rotatebox{90}{\scriptsize{~~~~~~~~~~~~~~~~~~~~\normalsize{Weather}}}
    \subfigure{
        \begin{minipage}[t]{0.44\linewidth}
            \centering
            \includegraphics[width=1\linewidth]{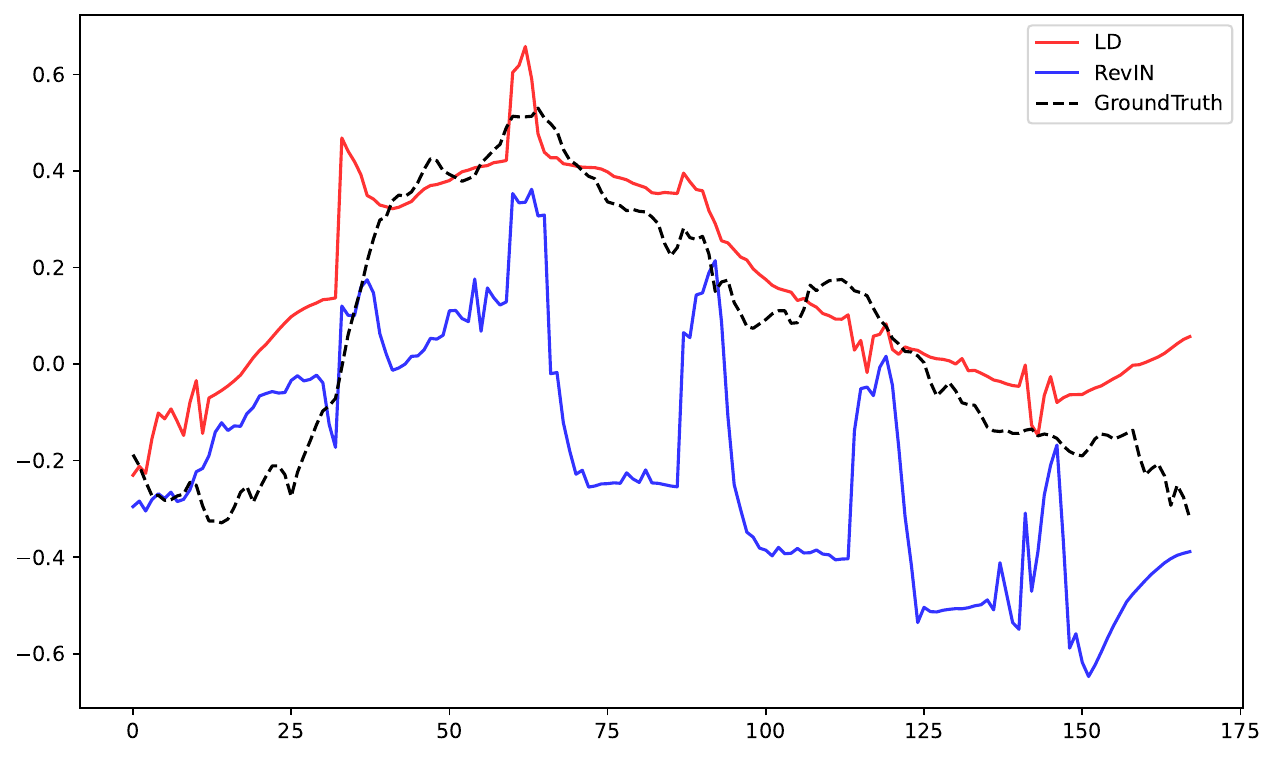}
        \end{minipage}
    }
    \subfigure{
        \begin{minipage}[t]{0.44\linewidth}
            \centering
            \includegraphics[width=1\linewidth]{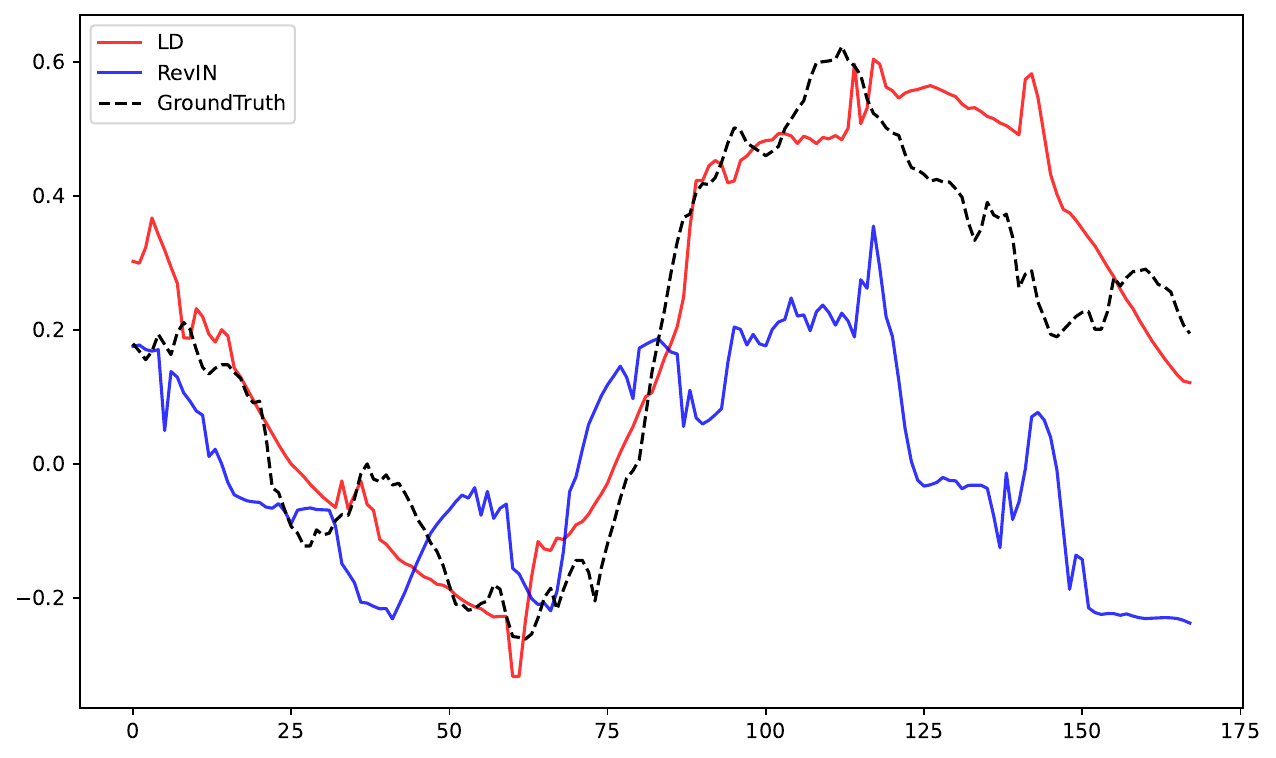}
        \end{minipage}
    }
    
    \setcounter{subfigure}{0}
    \hspace{-1.3mm}
     \rotatebox{90}{\scriptsize{~~~~~~~~~~~~~~~~~~~~\normalsize{Exchange}}}
    \subfigure{
        \begin{minipage}[t]{0.44\linewidth}
            \centering
            \includegraphics[width=1\linewidth]{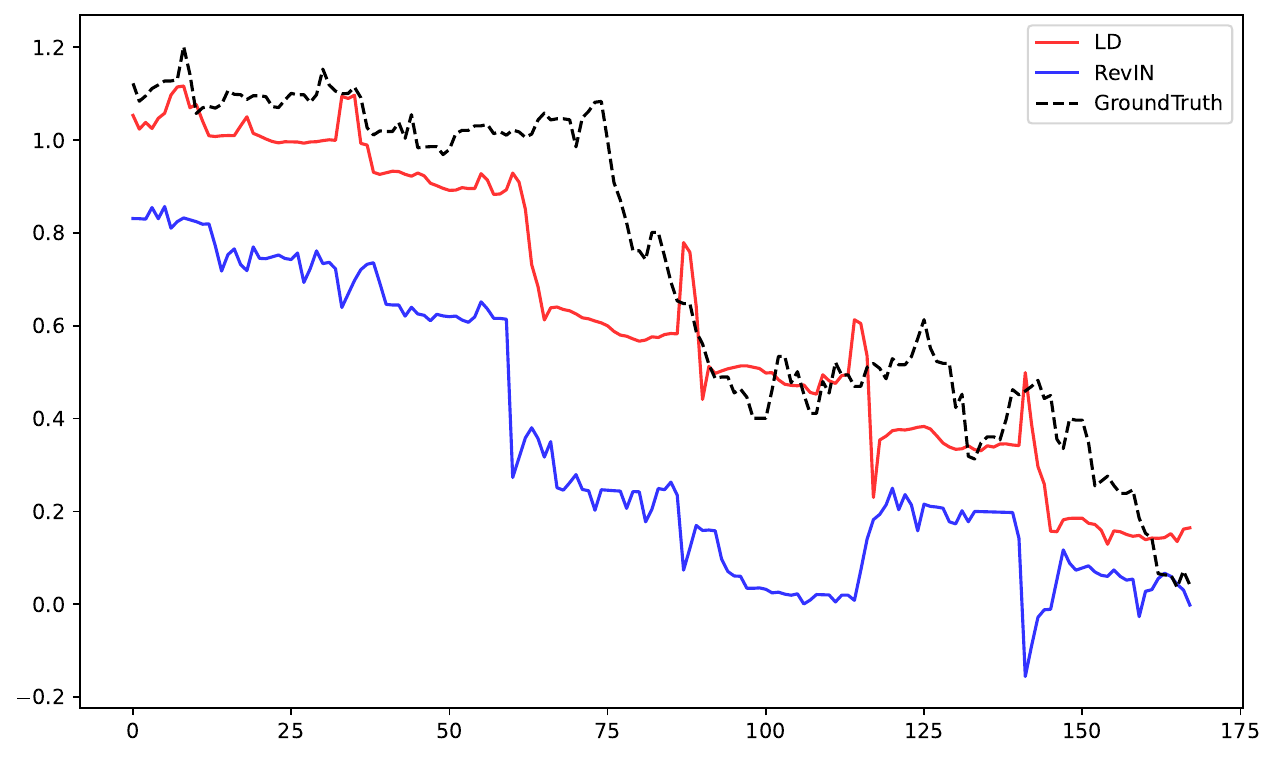}
        \end{minipage}
    }
    \hspace{-2.0mm}
    \subfigure{
        \begin{minipage}[t]{0.445\linewidth}
            \centering
            \includegraphics[width=1\linewidth]{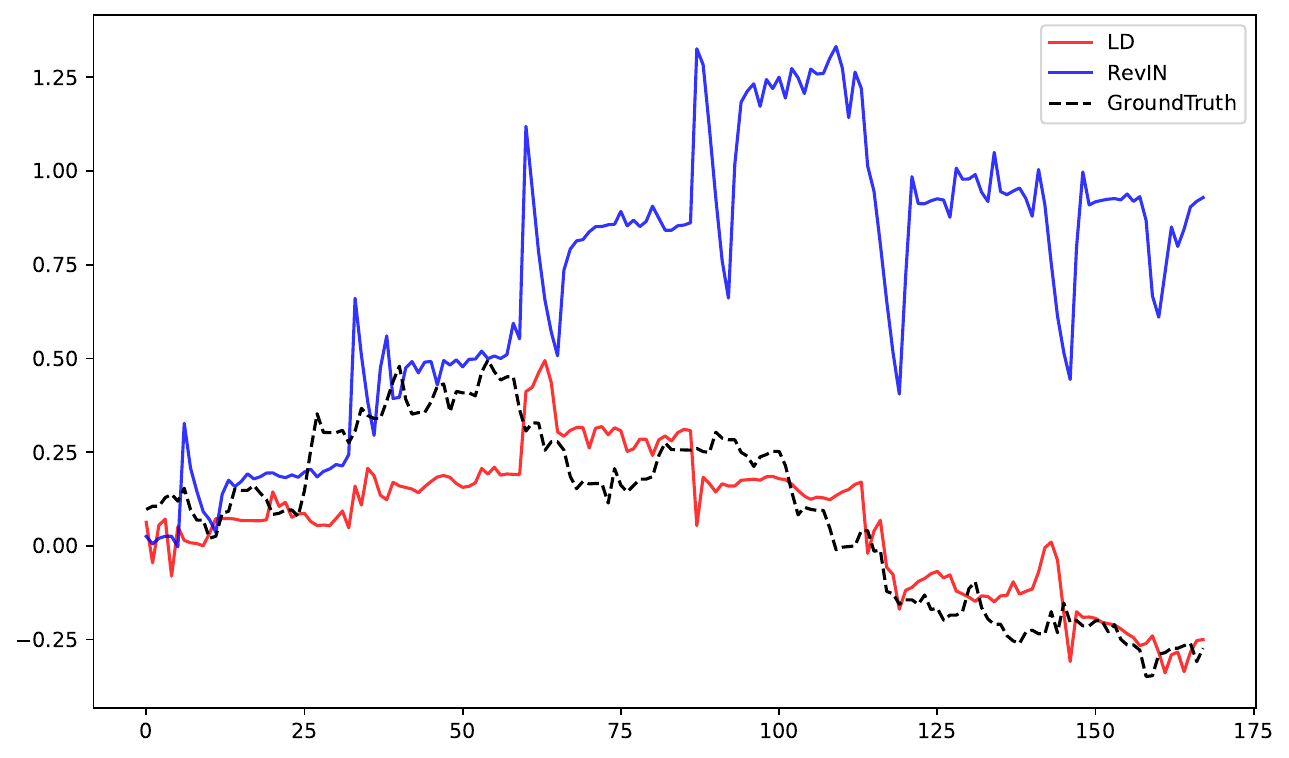}
        \end{minipage}
    }

    \caption{Visualization of several 168-long prediction results for LD and RevIN (Informer as backbone model).}
    \label{pred-ground}
\end{figure*}

\begin{figure*}[tb]\rmfamily
    \centering
    \subfigure{
        \rotatebox{90}{\scriptsize{~~~~~~~~~~~\normalsize{ETTh1}}}
        \hspace{1mm}
	\begin{minipage}[t]{0.23\linewidth}
		\centering
		\includegraphics[width=1\linewidth]{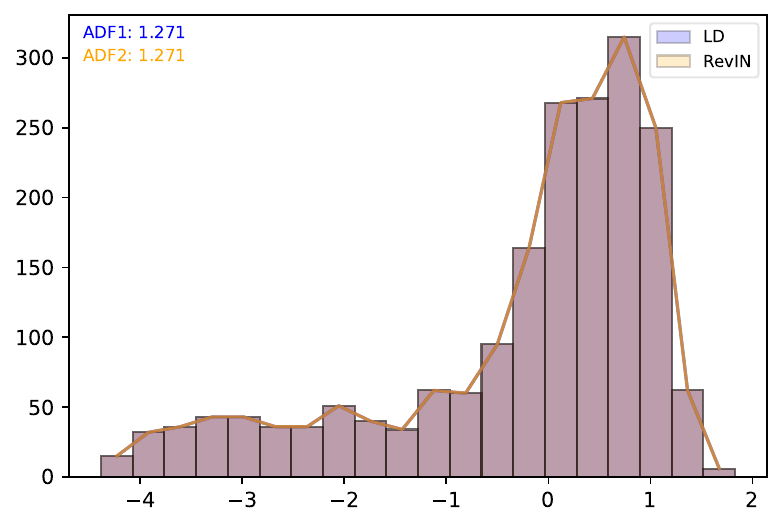}
	\end{minipage}
        \hspace{-3mm}
    }
    \subfigure{
	\begin{minipage}[t]{0.23\linewidth}
		\centering
		\includegraphics[width=1\linewidth]{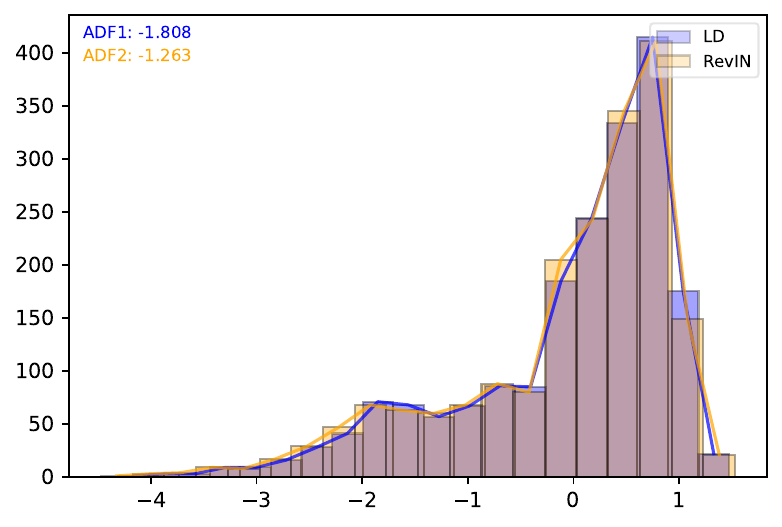}
	\end{minipage}
        \hspace{-3mm}
    }
    \subfigure{
	\begin{minipage}[t]{0.23\linewidth}
		\centering
		\includegraphics[width=1\linewidth]{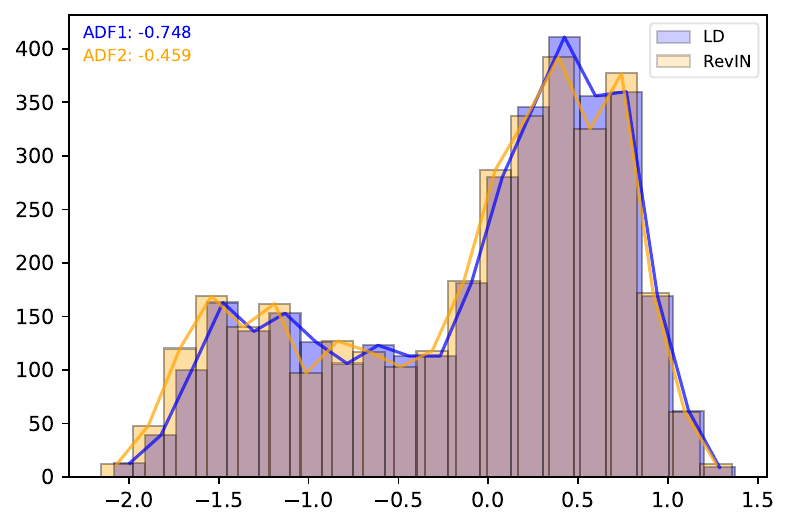}
	\end{minipage}
        \hspace{-3mm}
    }
    \subfigure{
	\begin{minipage}[t]{0.23\linewidth}
		\centering
		\includegraphics[width=1\linewidth]{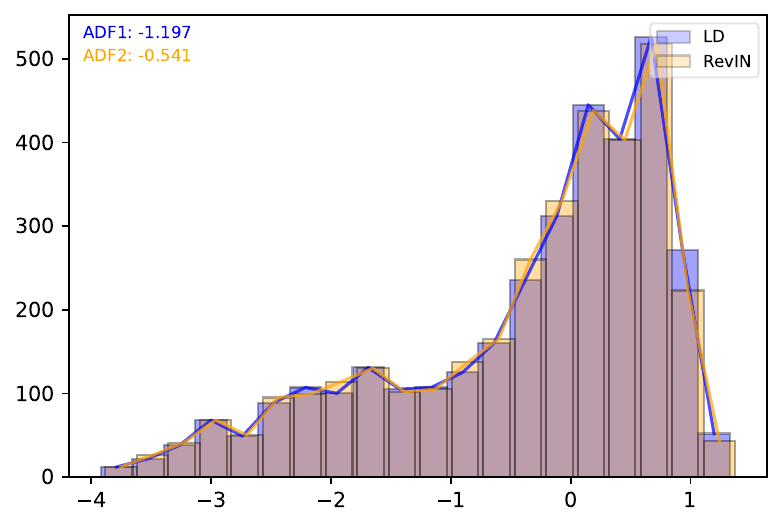}
	\end{minipage}
    } 
    
    \setcounter{subfigure}{0}
    \subfigure{
        \rotatebox{90}{\scriptsize{~~~~~~~~~~\normalsize{Weather}}}
        \hspace{1mm}
	\begin{minipage}[t]{0.23\linewidth}
		\centering
		\includegraphics[width=1\linewidth]{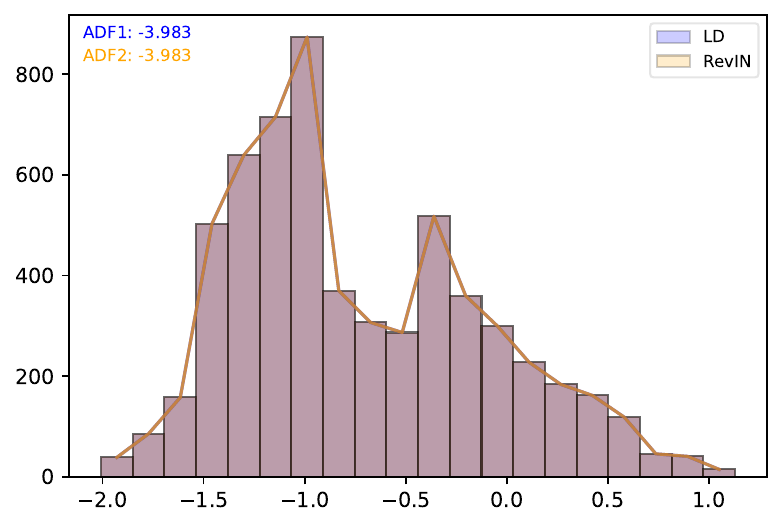}
	\end{minipage}
        \hspace{-3mm}
    }
    \subfigure{
	\begin{minipage}[t]{0.23\linewidth}
		\centering
		\includegraphics[width=1\linewidth]{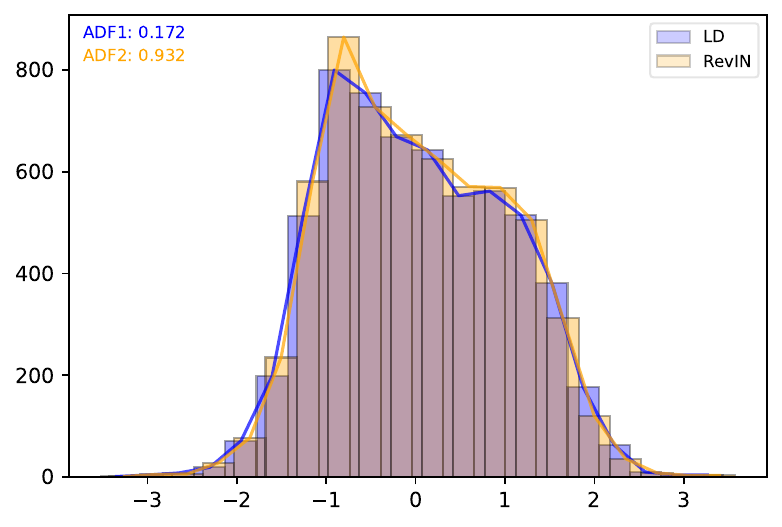}
	\end{minipage}
        \hspace{-3mm}
    }
    \subfigure{
	\begin{minipage}[t]{0.23\linewidth}
		\centering
		\includegraphics[width=1\linewidth]{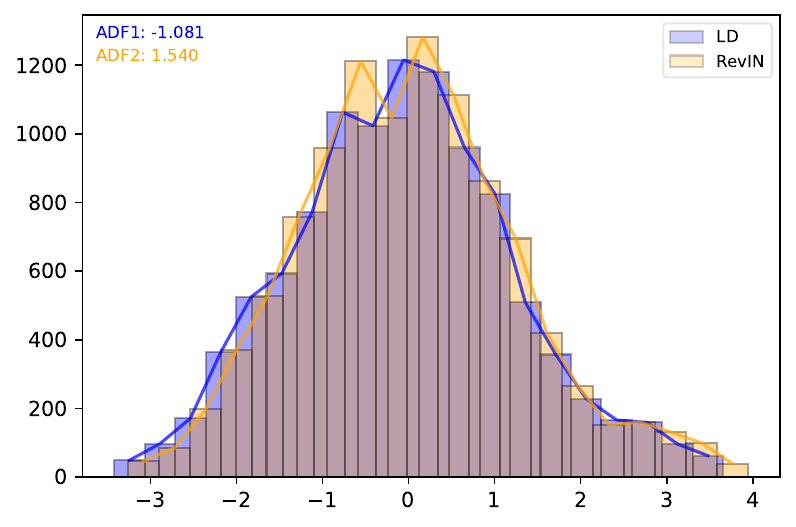}
	\end{minipage}
        \hspace{-3mm}
    }
    \subfigure{
	\begin{minipage}[t]{0.23\linewidth}
		\centering
		\includegraphics[width=1\linewidth]{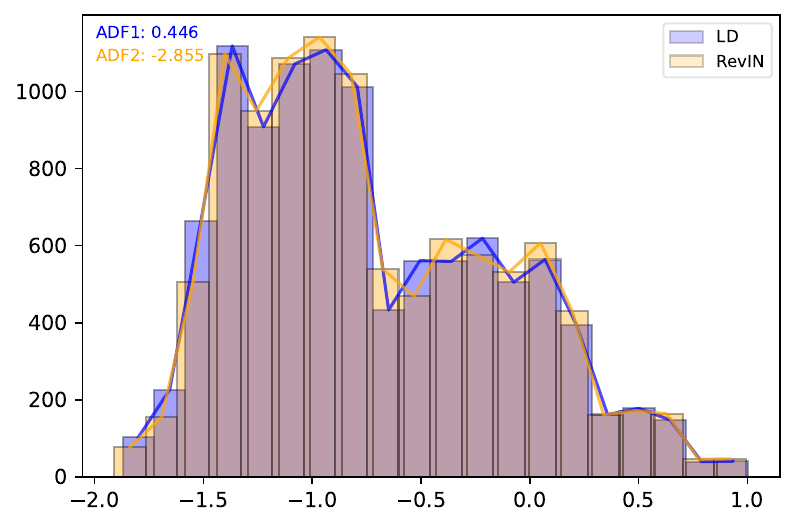}
	\end{minipage}
    } 
    
    \setcounter{subfigure}{0}
    \subfigure[Original Input]{
        \rotatebox{90}{\scriptsize{~~~~~~~~~\normalsize{Exchange}}}
        \hspace{1mm}
	\begin{minipage}[t]{0.23\linewidth}
		\centering
		\includegraphics[width=1\linewidth]{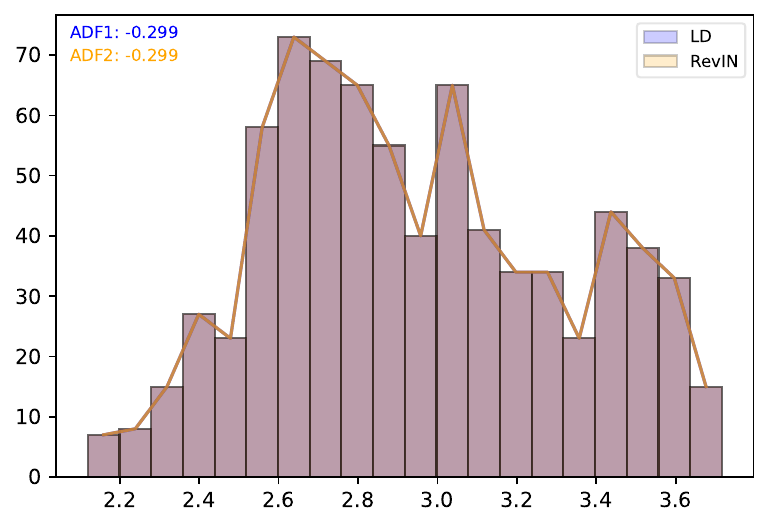}
	\end{minipage}
        \hspace{-3mm}
	}
    \subfigure[Normalized]{
	\begin{minipage}[t]{0.23\linewidth}
		\centering
		\includegraphics[width=1\linewidth]{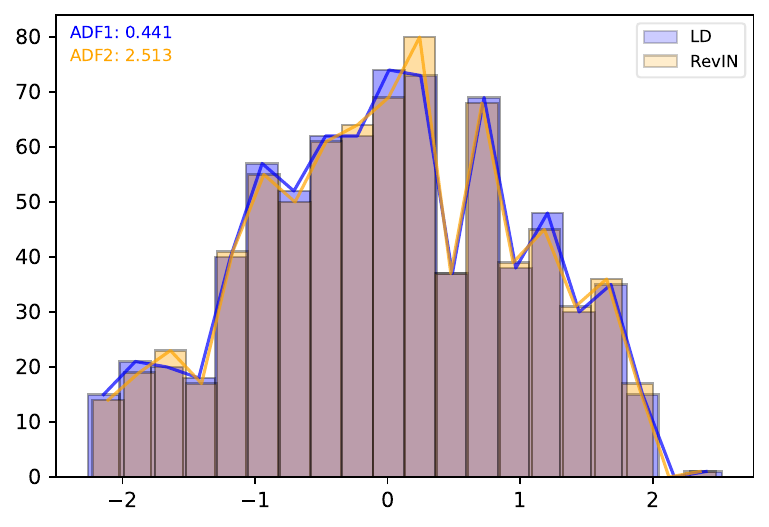}
	\end{minipage}
        \hspace{-3mm}
    }
    \subfigure[Model Output]{
	\begin{minipage}[t]{0.23\linewidth}
		\centering
		\includegraphics[width=1\linewidth]{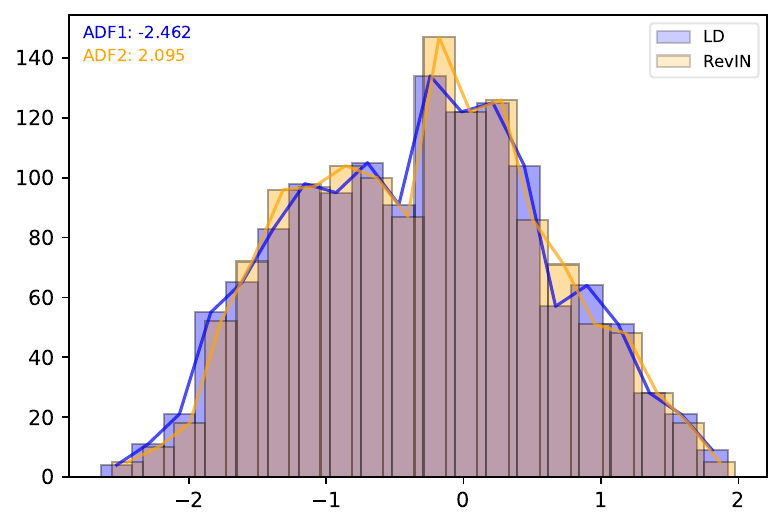}
	\end{minipage}
        \hspace{-3mm}
    }
    \subfigure[Denormalized]{
	\begin{minipage}[t]{0.23\linewidth}
		\centering
		\includegraphics[width=1\linewidth]{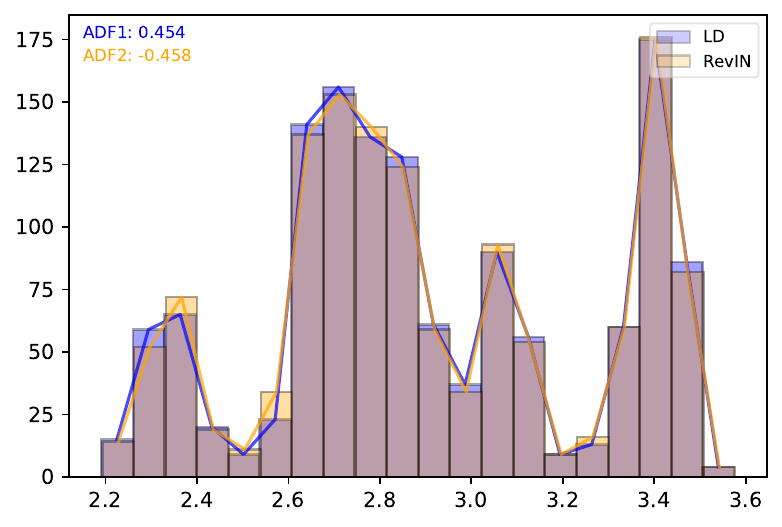}
	\end{minipage}
    }
    \vspace{4mm}
    \caption{Effectiveness of LD in turning non-stationary test data into stationary data compared with RevIN. From left to right, the distributions of the following data are shown: test data, test data normalized by LCD and RevIN, backbone model output, and the final prediction after denormalization (SCINet as backbone model).}
    \label{in-out-distribution}
\end{figure*}

\textbf{Comparison with the SOTA model RevIN.} RevIN is recognized as the current leading normalization model, relying solely on learnable parameters to approximate the distribution. However, it fails to account for shifts occurring within individual instances. As illustrated in \ref{LDwo-tab3}, the performance of our LD model consistently surpasses that of RevIN across all models and datasets. Notably, LD demonstrates a significant advantage over RevIN in the Informer model, while it achieves a modest improvement in both the N-BEATS and SCINet models. The primary strength of LD lies in its ability to address inner-instance distribution shifts by adapting the distribution at each time step. The results presented in \ref{pred-ground} further indicate that LD is more adept at accommodating changes in inner-instance distributions. In the initial phase, both models accurately predict the outcomes. However, as the inner-instance distribution shifts occur, RevIN struggles to align with the ground truth series, while LD maintains accurate predictions. Additionally, \ref{in-out-distribution} reveals that normalizing data with LD results in smaller ADF values for both input and output of the backbone model, suggesting stationarity. One could also find that the normalized inputs and outputs of the backbone models are more compliant with the normal distribution than the original inputs and final outputs, which shows the effectiveness of the two normalization methods.

\begin{table*}[htb]\rmfamily
    \caption{The performance improvements of LCD on three models—DLinear, PatchTST, and iTransformer across 6 datasets. The prediction length is $96$.}
\label{lcd-imp}
    \vskip 0.15in
    \setlength{\tabcolsep}{2pt}
    \begin{footnotesize}
    \centering
    \begin{tabular}{c|cccccc|cccccc|cccccc}
        \toprule
        \multirow{2}{*}{Dataset} 
        & \multicolumn{2}{c}{DLinear} & \multicolumn{2}{c}{+ LCD-linear}  & \multicolumn{2}{c|}{+ LCD-as} & \multicolumn{2}{c}{PatchTST} & \multicolumn{2}{c}{+ LCD-linear} & \multicolumn{2}{c|}{+ LCD-as} & \multicolumn{2}{c}{iTransformer} & \multicolumn{2}{c}{+ LCD-linear} & \multicolumn{2}{c}{+ LCD-as} \\
        & MSE & MAE & MSE & MAE & MSE & MAE & MSE & MAE & MSE & MAE & MSE & MAE & MSE & MAE & MSE & MAE & MSE & MAE \\
        \midrule
        Weather & 0.160 & 0.265 & 0.147 & 0.252 & \textbf{0.145} & \textbf{0.248} & 0.145 & \textbf{0.237} & \textbf{0.139} & 0.242 & 0.142 & 0.247 & 0.168 & 0.260 & \textbf{0.139} & \textbf{0.242} & 0.147 & 0.249 \\
        Exchange & 0.087 & 0.213 & 0.082 & 0.209 & \textbf{0.079} & \textbf{0.207} & 0.108 & 0.241 & 0.087 & 0.215 & \textbf{0.081} & \textbf{0.208} & 0.096 & 0.219 & \textbf{0.082} & \textbf{0.209} & 0.089 & 0.225 \\
        ETTh1 & 0.451 & 0.447 & 0.441 & \textbf{0.445} & \textbf{0.434} & 0.451 & 0.446 & 0.460 & 0.433 & 0.453 & \textbf{0.425} & \textbf{0.452} & 0.479 & 0.475 & \textbf{0.441} & \textbf{0.445} & 0.444 & 0.460 \\
        ETTh2 & 0.191 & 0.306 & 0.187 & 0.295 & \textbf{0.184} & \textbf{0.291} & 0.215 & 0.326 & 0.194 & 0.298 & \textbf{0.180} & \textbf{0.290} & 0.195 & 0.301 & \textbf{0.186} & \textbf{0.291} & 0.187 & 0.295 \\
        Traffic-s & 0.600 & 0.452 & \textbf{0.529} & 0.424 & 0.535 & \textbf{0.423} & 0.356 & 0.309 & 0.345 & 0.302 & \textbf{0.344} & \textbf{0.301} & 0.325 & 0.299 & \textbf{0.318} & 0.288 & 0.320 & \textbf{0.285} \\
        ECL-s & 0.336 & 0.388 & \textbf{0.312} & \textbf{0.378} & 0.320 & 0.381 & 0.277 & 0.351 & \textbf{0.273} & \textbf{0.347} & 0.275 & 0.350 & 0.240 & 0.317 & 0.231 & 0.316 & \textbf{0.229} & \textbf{0.314} \\
        Imp. & & & 6.29\% & 3.27\% & 6.93\% & 3.58\% & & & 6.80\% & 3.70\% & 8.69\% & 4.19\% & & & 8.38\% & 4.23\% & 6.17\% & 2.20\% \\
        \bottomrule
    \end{tabular}
    \end{footnotesize}
\vskip -0.1in
\end{table*}

\begin{table*}[htb]\rmfamily
    \caption{Comparison of LCD and baseline models SAN, Dish-TS, and NST across 5 datasets, using FEDformer and Autoformer as backbone models. The top results are shown in bold.}
    
    \label{Table5}
    \centering
    \setlength{\tabcolsep}{3pt}
    \begin{footnotesize}
    \begin{tabular}{c|c|cccccccc|cccccccc}
        \toprule
        \multirow{3}{*}{\rotatebox{90}{Dataset}}
        & Model & \multicolumn{8}{c|}{FEDformer}  & \multicolumn{8}{c}{Autoformer} \\  
        & Method & \multicolumn{2}{c}{+ LCD-linear} & \multicolumn{2}{c}{+ SAN} & \multicolumn{2}{c}{+ Dish-TS} & \multicolumn{2}{c|}{+ NST} & \multicolumn{2}{c}{+ LCD-linear} & \multicolumn{2}{c}{+ SAN} & \multicolumn{2}{c}{+ Dish-TS} & \multicolumn{2}{c}{+ NST}\\
        & Metric & MSE & MAE & MSE & MAE & MSE & MAE & MSE & MAE & MSE & MAE & MSE & MAE & MSE & MAE & MSE & MAE \\
        \midrule
        \multirow{4}{*}{\rotatebox{90}{Weather}} 
        & 96  & \textbf{0.143} &\textbf{0.246} & 0.146 & 0.243 & 0.156 & 0.260 & 0.255 & 0.364 & \textbf{0.140} & \textbf{0.243} & 0.162 & 0.258 & 0.171 & 0.281 & 0.192 & 0.287 \\
        & 192   & \textbf{0.182} &\textbf{0.286} & 0.201 & 0.298 & 0.233 & 0.328 & 0.345 & 0.403 & \textbf{0.182} & \textbf{0.283} & 0.208 & 0.305 & 0.221 & 0.313 & 0.243 & 0.324 \\
        & 336  & \textbf{0.255} &\textbf{0.342} & 0.284 & 0.353 & 0.281 & 0.358 & 0.408 & 0.453 & \textbf{0.262} & \textbf{0.344} & 0.273 & 0.356 & 0.333 & 0.404 & 0.306 & 0.368 \\
        & 720  & \textbf{0.332} &\textbf{0.396} & 0.372 & 0.409 & 0.360 & 0.412 & 0.429 & 0.461 & \textbf{0.328} & \textbf{0.393} & 0.350 & 0.409 & 0.383 & 0.440 & 0.396 & 0.424 \\
         \midrule
        \multirow{4}{*}{\rotatebox{90}{Exchange}} 
        & 48   & \textbf{0.042} &\textbf{0.145} & 0.057 & 0.169 & 0.056 & 0.169 & 0.071 & 0.192 & \textbf{0.044} & \textbf{0.145} & 0.047 & 0.154 & 0.071 & 0.191 & 0.152 & 0.284 \\
        & 96  & \textbf{0.087} &\textbf{0.218} & 0.097 & 0.222 & 0.262 & 0.362 & 0.139 & 0.268 & \textbf{0.090} & \textbf{0.217} & 0.092 & 0.219 & 0.240 & 0.371 & 0.310 & 0.411 \\
        & 192   & \textbf{0.188} &\textbf{0.323} & 0.198 & 0.325 & 0.908 & 0.677 & 0.295 & 0.396 & \textbf{0.180} & \textbf{0.322} & 0.192 & 0.327 & 0.389 & 0.444 & 0.589 & 0.578 \\
        & 336  & \textbf{0.360} &\textbf{0.465} & 0.570 & 0.540 & 0.602 & 0.625 & 0.535 & 0.547 & \textbf{0.375} & \textbf{0.475} & 0.425 & 0.492 & 0.880 & 0.641 & 1.274 & 0.862 \\
        \midrule
        \multirow{4}{*}{\rotatebox{90}{Electricity}} 
        & 96  & \textbf{0.185} &\textbf{0.281} & 0.195 & 0.301 & 0.200 & 0.305 & 0.198 & 0.310 & \textbf{0.184} & \textbf{0.281} & 0.211 & 0.314 & 0.201 & 0.309 & 0.196 & 0.299 \\
        & 192   & \textbf{0.184} &\textbf{0.285} & 0.204 & 0.307 & 0.206 & 0.313 & 0.204 & 0.316 & \textbf{0.171} & \textbf{0.272} & 0.239 & 0.336 & 0.217 & 0.321 & 0.214 & 0.317 \\
        & 336  & \textbf{0.200} &\textbf{0.300} & 0.215 & 0.317 & 0.231 & 0.336 & 0.220 & 0.335 & \textbf{0.176} & \textbf{0.283} & 0.233 & 0.334 & 0.222 & 0.325 & 0.232 & 0.334 \\
        & 720  & \textbf{0.213} &\textbf{0.312} & 0.214 & 0.319 & 0.272 & 0.362 & 0.246 & 0.357 & \textbf{0.209} & \textbf{0.312} & 0.312 & 0.403 & 0.256 & 0.356 & 0.231 & 0.333 \\
        \midrule
        \multirow{4}{*}{\rotatebox{90}{Traffic}} 
        & 96  & 0.616 & 0.359 & \textbf{0.534} &\textbf{0.332} & 0.662 & 0.375 & 0.607 & 0.387 & \textbf{0.578} & \textbf{0.371} & 0.619 & 0.398 & 0.709 & 0.401 & 0.721 & 0.415 \\
        & 192   & \textbf{0.554} &\textbf{0.326} & 0.567 & 0.348 & 0.648 & 0.360 & 0.601 & 0.376 & \textbf{0.555} & \textbf{0.331} & 0.606 & 0.368 & 0.669 & 0.365 & 0.611 & 0.382 \\
        & 336  & \textbf{0.548} &\textbf{0.328} & 0.596 & 0.356 & 0.657 & 0.3364 & 0.644 & 0.402 & \textbf{0.576} & \textbf{0.343} & 0.595 & 0.364 & 0.688 & 0.380 & 0.671 & 0.377 \\
        & 720  & \textbf{0.598} &\textbf{0.349} & 0.616 & 0.367 & 0.731 & 0.397 & 0.634 & 0.383  & 0.753 & 0.398 & \textbf{0.629} & 0.380 & 0.739 & 0.398 & 0.677 & \textbf{0.378} \\
        \midrule
        \multirow{4}{*}{\rotatebox{90}{ETTh2}} 
        & 96  & \textbf{0.191} & \textbf{0.296} & 0.218 & 0.314 & 0.259 & 0.354 & 0.203 & 0.309 & \textbf{0.183} &\textbf{0.290} & 0.188 & 0.296 & 0.227 & 0.332  & 0.228 & 0.329 \\
        & 192 & \textbf{0.199} & \textbf{0.311} & 0.227 & 0.332 & 0.272 & 0.374  & 0.234 & 0.335 & \textbf{0.196} &\textbf{0.317} & 0.222 & 0.328 & 0.280 & 0.371 & 0.246 & 0.346 \\
        & 336  & \textbf{0.222} & \textbf{0.333} & 0.265 & 0.368 & 0.349 & 0.426 & 0.258 & 0.362 & \textbf{0.213} &\textbf{0.328} & 0.232 & 0.340 & 0.303 & 0.392 & 0.278 & 0.377 \\
        & 720 & 0.325 & \textbf{0.403} & \textbf{0.309} & 0.405 & 0.619 & 0.558  & 0.338 & 0.437 & \textbf{0.224} &\textbf{0.338} & 0.293 & 0.390 & 0.401 & 0.470 & 0.319 & 0.409 \\
        \bottomrule
    \end{tabular}
    \end{footnotesize}
\vskip -0.1in
\end{table*}

\begin{figure*}[tb]\rmfamily
    \centering
    \subfigure{
        \rotatebox{90}{\scriptsize{~~~~~~~~~~~~~~~~\normalsize{ETTh2}}}
        \hspace{1mm}
	\begin{minipage}[t]{0.31\linewidth}
		\centering
		\includegraphics[width=1\linewidth]{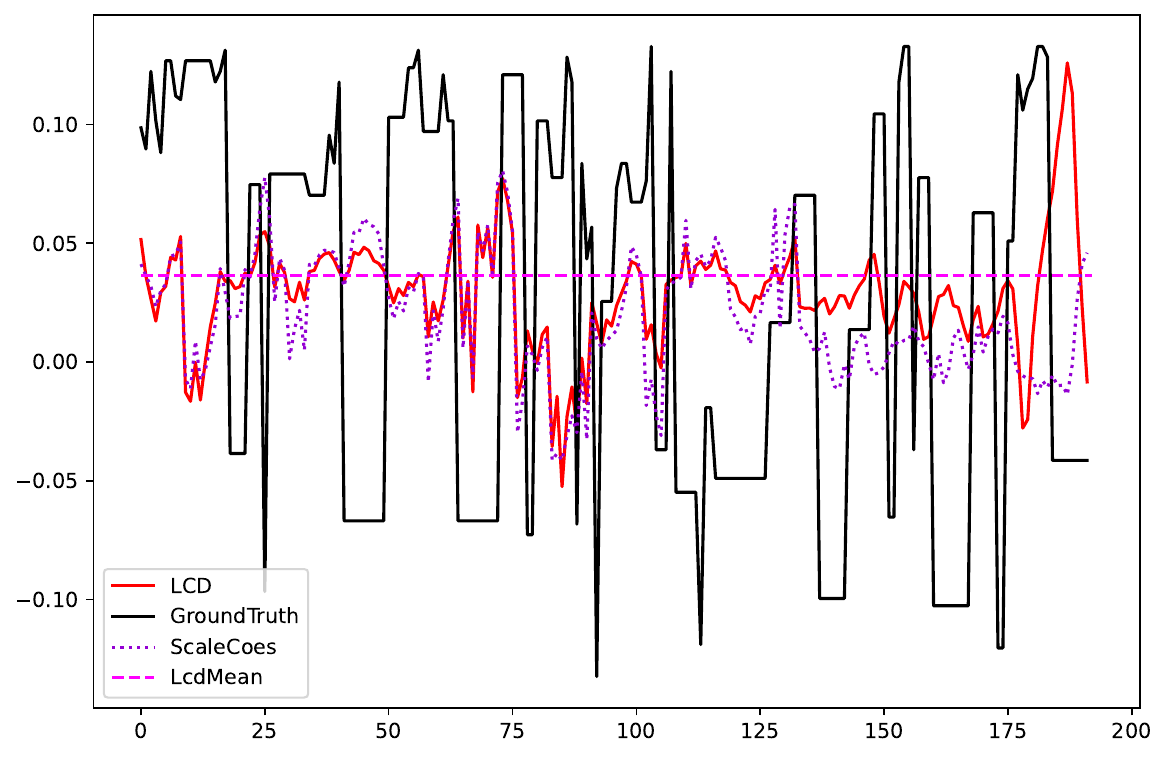}
	\end{minipage}
        \hspace{-3mm}
    }
    \subfigure{
	\begin{minipage}[t]{0.31\linewidth}
		\centering
		\includegraphics[width=1\linewidth]{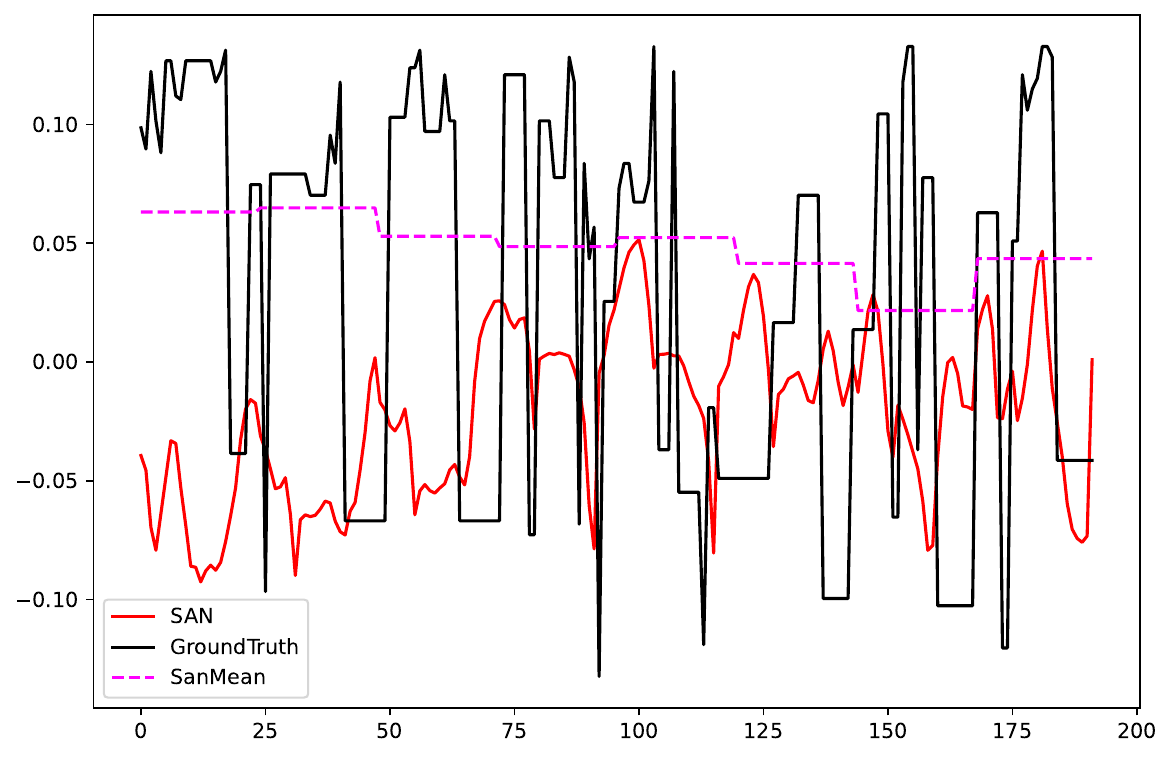}
	\end{minipage}
        \hspace{-3mm}
    }
    \subfigure{
	\begin{minipage}[t]{0.31\linewidth}
		\centering
		\includegraphics[width=1\linewidth]{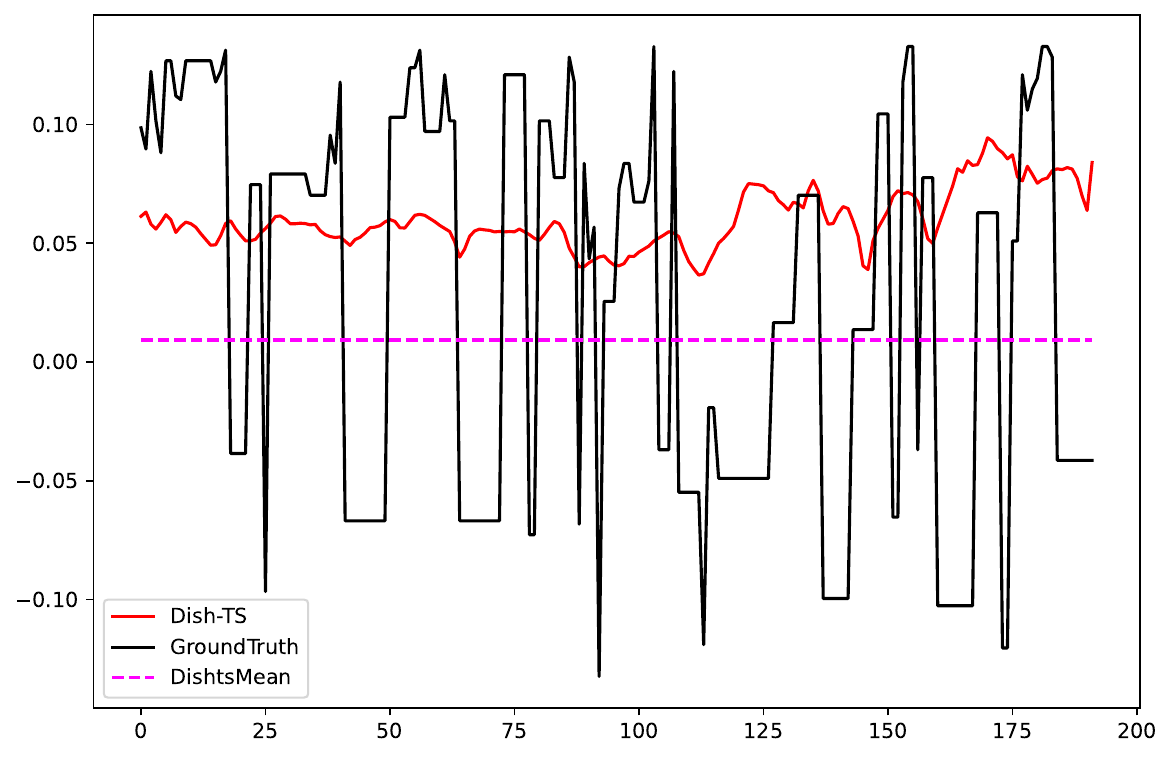}
	\end{minipage}
    }
    
    \setcounter{subfigure}{0}
    \subfigure{
        \rotatebox{90}{\scriptsize{~~~~~~~~~~~~~~~~\normalsize{Weather}}}
        \hspace{1mm}
	\begin{minipage}[t]{0.31\linewidth}
		\centering
		\includegraphics[width=1\linewidth]{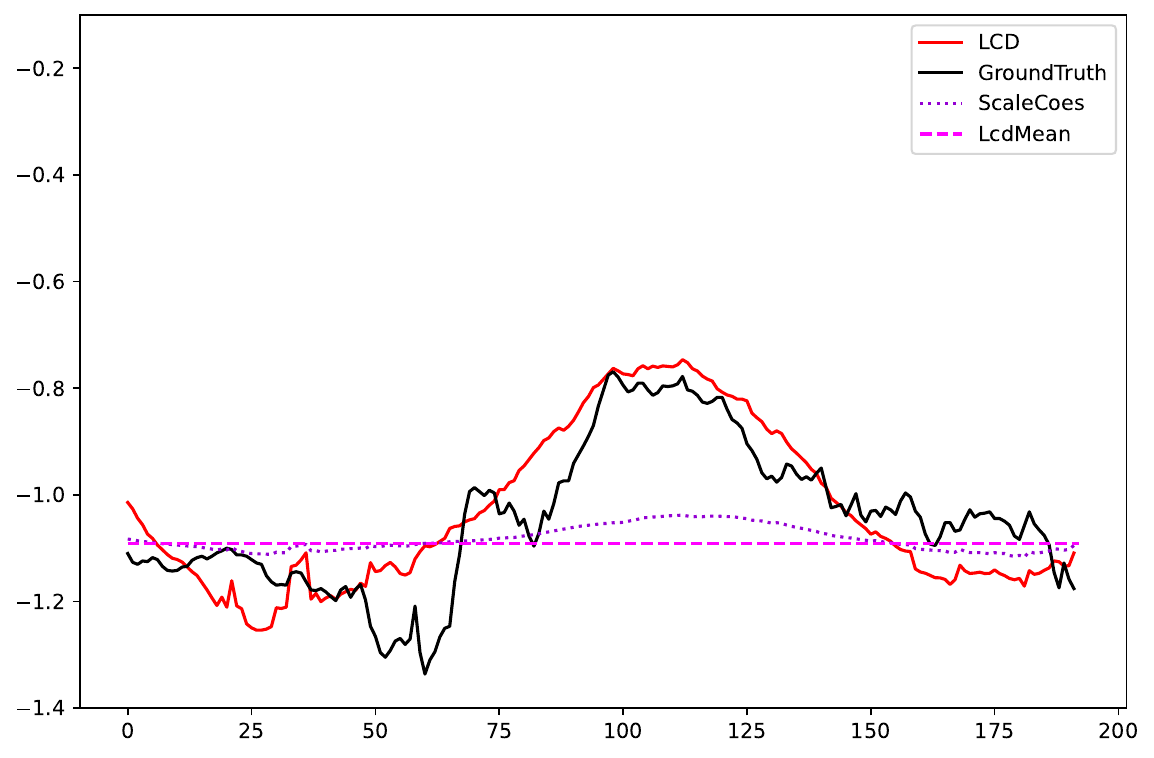}
	\end{minipage}
        \hspace{-3mm}
    }
    \subfigure{
	\begin{minipage}[t]{0.31\linewidth}
		\centering
		\includegraphics[width=1\linewidth]{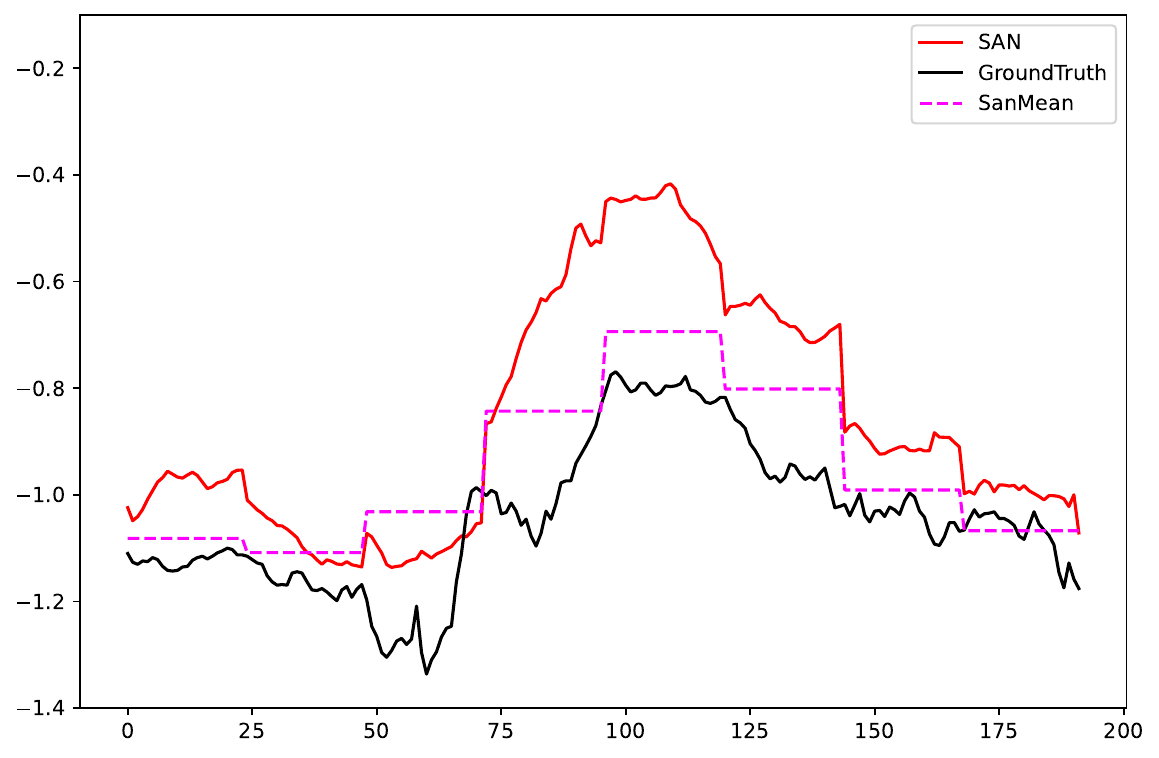}
	\end{minipage}
        \hspace{-3mm}
    }
    \subfigure{
	\begin{minipage}[t]{0.31\linewidth}
		\centering
		\includegraphics[width=1\linewidth]{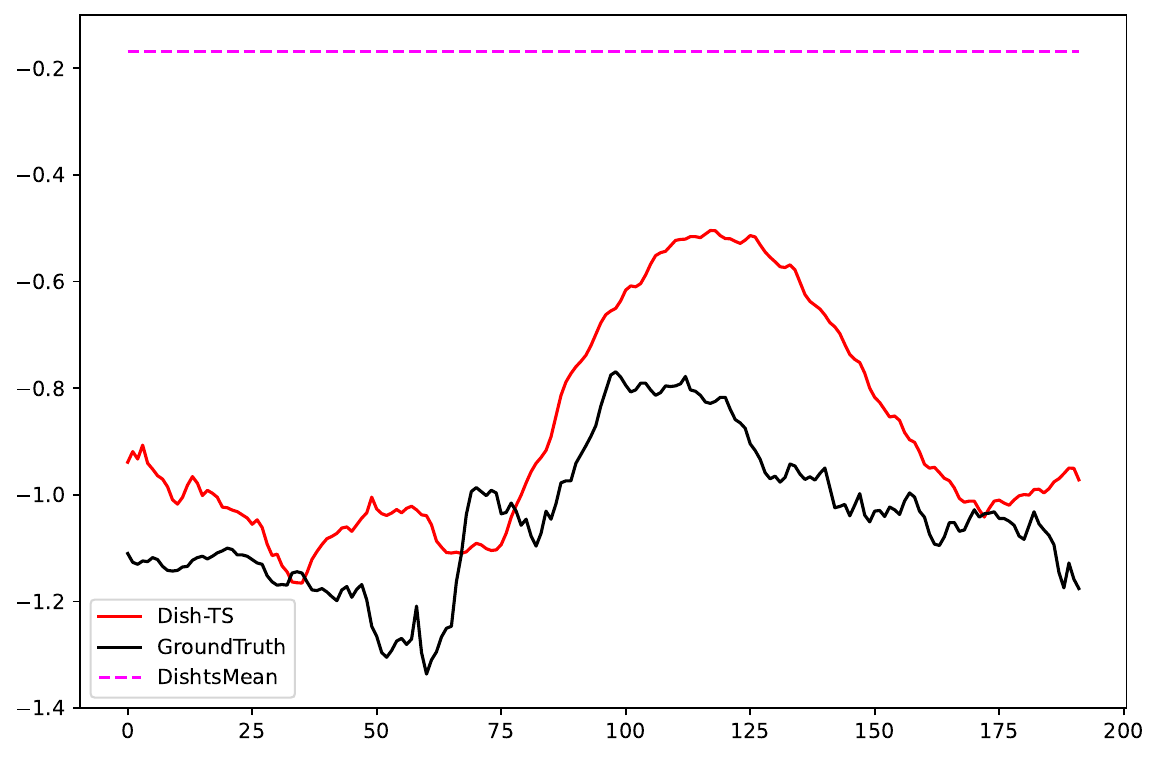}
	\end{minipage}
    }
    
    \setcounter{subfigure}{0}
    \subfigure[LCD-linear]{
        \rotatebox{90}{\scriptsize{~~~~~~~~~~~~~~~\normalsize{Exchange}}}
        \hspace{1mm}
	\begin{minipage}[t]{0.31\linewidth}
		\centering
		\includegraphics[width=1\linewidth]{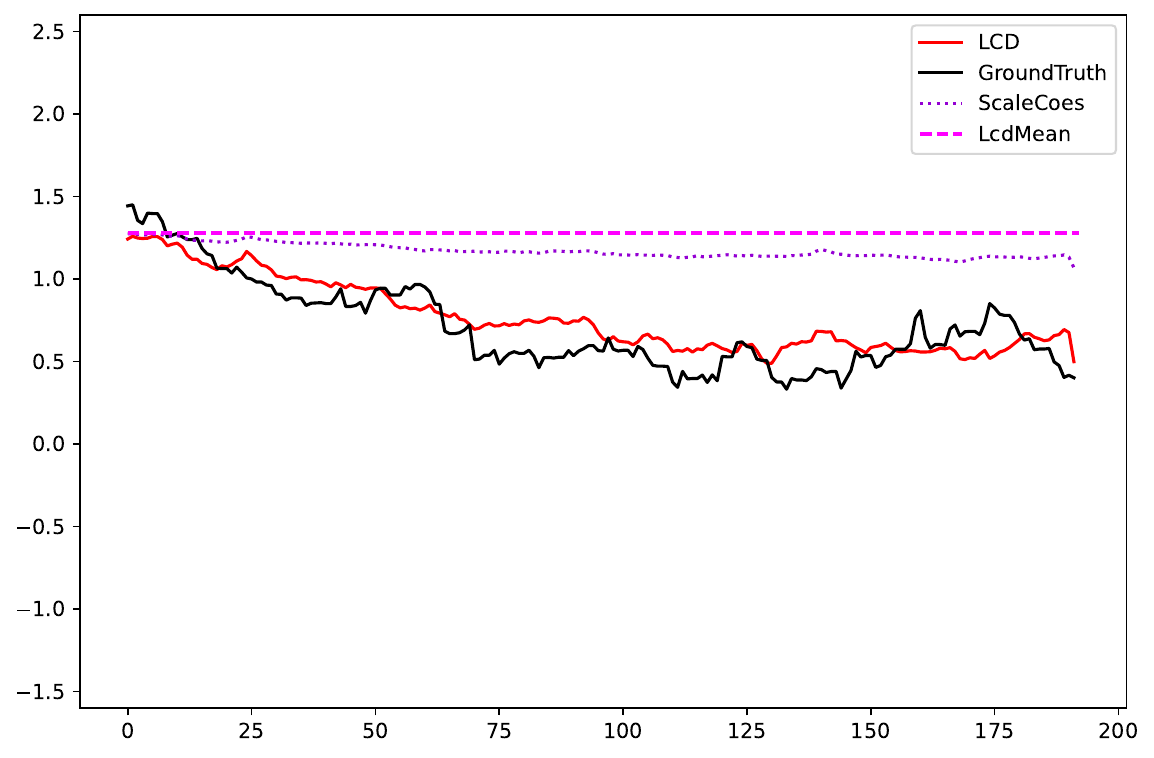}
	\end{minipage}
        \hspace{-3mm}
	}
    \subfigure[SAN]{
	\begin{minipage}[t]{0.31\linewidth}
		\centering
		\includegraphics[width=1\linewidth]{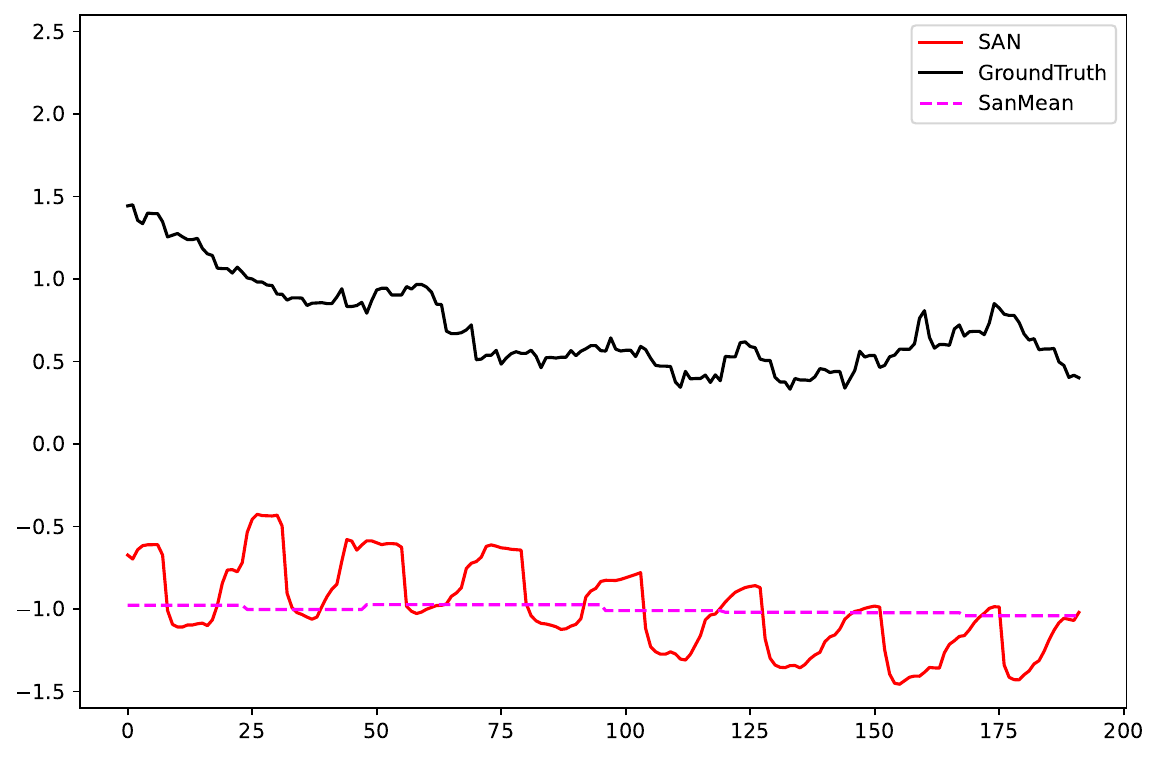}
	\end{minipage}
        \hspace{-3mm}
    }
    \subfigure[Dish-TS]{
	\begin{minipage}[t]{0.31\linewidth}
		\centering
		\includegraphics[width=1\linewidth]{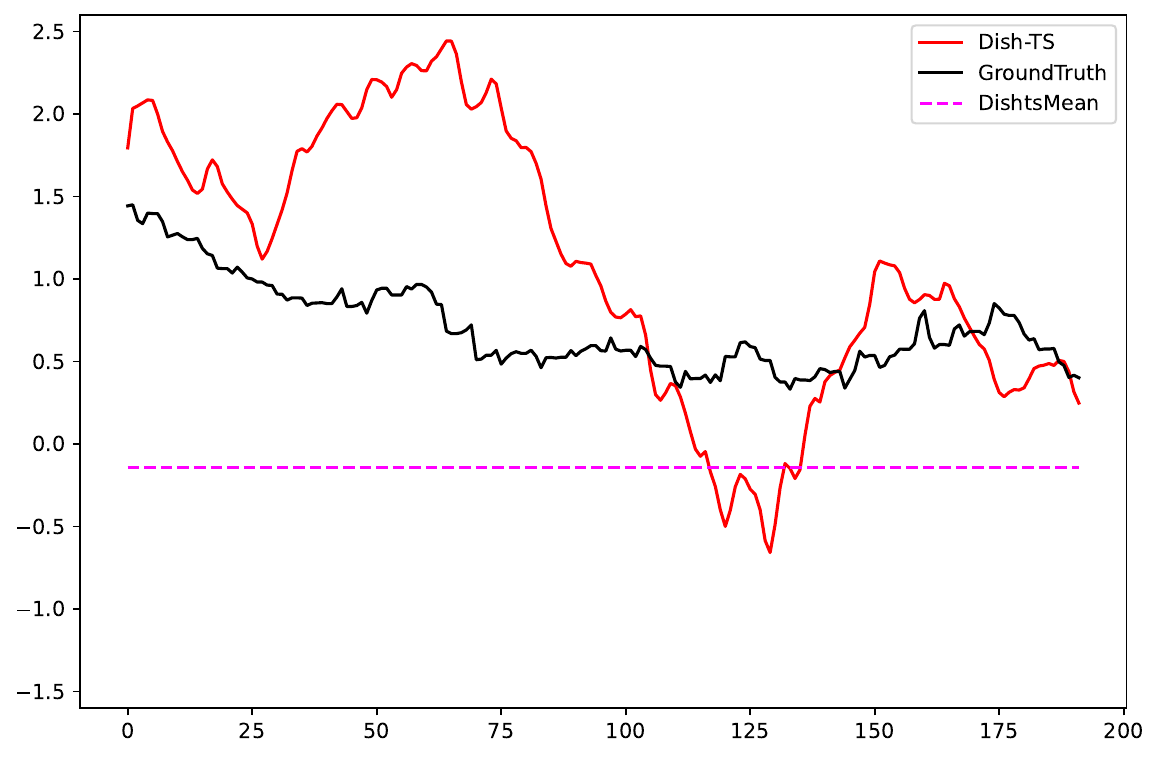}
	\end{minipage}}
    \caption{Visualization of examples of 192-long prediction results (Autoformer as backbone model). ScaleCoes are scaling coefficients $\bm{S}$.}
    \label{fig-lcd-san-dishts}
\end{figure*}

\begin{table*}[htbp]\rmfamily
    \caption{Comparison between LCD-linear and Instance-level LCD-linear. Instance-level LCD-linear scales different time steps with the same scaling coefficients. The horizon window is set to $96$.} 
    \label{Table8}
    \centering
    \begin{tabular}{c|cccc|cccc|cccc}
        \toprule
        \multirow{3}{*}{Dataset} 
        & \multicolumn{4}{c|}{DLinear + LCD-linear} & \multicolumn{4}{c|}{PatchTST + LCD-linear} & \multicolumn{4}{c}{iTransformer + LCD-linear} \\
        & \multicolumn{2}{c}{Instance-level} & \multicolumn{2}{c|}{Point-level} & \multicolumn{2}{c}{+ Instance-level} & \multicolumn{2}{c|}{Point-level} & \multicolumn{2}{c}{Instance-level} & \multicolumn{2}{c}{Point-level} \\
        & MSE & MAE & MSE & MAE & MSE & MAE & MSE & MAE & MSE & MAE & MSE & MAE \\
        \midrule
        Weather & \textbf{0.139} & \textbf{0.243} & 0.147 & 0.252 & 0.143 & 0.246 & \textbf{0.139} & \textbf{0.242} & 0.140 & 0.243 & \textbf{0.139} & \textbf{0.242} \\
        Exchange & 0.087 & 0.223 & \textbf{0.082} & \textbf{0.209} & 0.089 & 0.221 & \textbf{0.087} & \textbf{0.215} & 0.084 & 0.213 & \textbf{0.082} & \textbf{0.209} \\
        ETTh1 & 0.698 & 0.601 & \textbf{0.441} & \textbf{0.445} & 0.518 & 0.509 & \textbf{0.433} & \textbf{0.453} & 0.494 & 0.481 & \textbf{0.441} & \textbf{0.445} \\
        ETTh2 & 0.217 & 0.324 & \textbf{0.187} & \textbf{0.295} & 0.209 & 0.312 & \textbf{0.194} & \textbf{0.298} & 0.202 & 0.312 & \textbf{0.186} & \textbf{0.291} \\
        Traffic-s & 0.609 & 0.459 & \textbf{0.529} & \textbf{0.424} & 0.375 & 0.323 & \textbf{0.345} & \textbf{0.302} & 0.373 & 0.327 & \textbf{0.318} & \textbf{0.288} \\
        ECL-s & 0.338 & 0.391 & \textbf{0.312} & \textbf{0.378} & 0.275 & 0.349 & \textbf{0.273} & \textbf{0.347} & 0.252 & 0.334 & \textbf{0.231} & \textbf{0.316} \\
        \bottomrule
    \end{tabular}
\end{table*}

\begin{table*}[tb]\rmfamily
    \caption{Comparison between LD and Instance-Level LD. Instance-level LD fits the shared mean of different time steps. The forecasting length is set to $168$.}
    \label{Table7}
    \centering
    \begin{tabular}{c|cccc|cccc|cccc}
        \toprule
        \multirow{3}{*}{Dataset} 
        & \multicolumn{4}{c|}{Informer + LD} & \multicolumn{4}{c|}{N-BEATS+ LD} & \multicolumn{4}{c}{SCINet+ LD} \\
        & \multicolumn{2}{c}{Instance-level} & \multicolumn{2}{c|}{Point-level} & \multicolumn{2}{c}{Instance-level} & \multicolumn{2}{c|}{Point-level} & \multicolumn{2}{c}{Instance-level} & \multicolumn{2}{c}{Point-level} \\
        & MSE & MAE & MSE & MAE & MSE & MAE & MSE & MAE & MSE & MAE & MSE & MAE \\
        \midrule
        Electricity & 0.199 & 0.301 & \textbf{0.196} & \textbf{0.299} & 0.179 & 0.273 & \textbf{0.178} & \textbf{0.272} & 0.164 & 0.264 & \textbf{0.162} & \textbf{0.262} \\
        Traffic & 0.650 & 0.354 & \textbf{0.636} & \textbf{0.345} & 0.517 & 0.339 & \textbf{0.510} & \textbf{0.337} & 0.488 & 0.333 & \textbf{0.486} & \textbf{0.332} \\
        ETTh1 & \textbf{0.597} & 0.566 & 0.607 & \textbf{0.564} & \textbf{0.493} & 0.499 & 0.498 & \textbf{0.498} & 0.495 & 0.486 & \textbf{0.493} & \textbf{0.485} \\
        ETTh2 & \textbf{0.255} & \textbf{0.356} & 0.276 & 0.378 & 0.205 & 0.317 & \textbf{0.202} & \textbf{0.313} & 0.207 & 0.314 & \textbf{0.207} & \textbf{0.313} \\
        Weather & \textbf{0.216} & 0.303 & 0.219 & \textbf{0.301} & 0.179 & 0.267 & \textbf{0.177} & \textbf{0.265} & \textbf{0.174} & \textbf{0.265} & 0.175 & 0.266 \\
        Exchange & \textbf{0.256} & \textbf{0.335} & 0.298 & 0.359 & 0.157 & 0.285 & \textbf{0.154} & \textbf{0.283} & 0.152 & 0.286 & \textbf{0.151} & \textbf{0.284} \\
        \bottomrule
    \end{tabular}
\end{table*}

\begin{table*}[tb]\rmfamily
    \caption{Comparison of using the same and different prediction functions $\bm{f}$ and $h$ for different features in LCD ($H = 96$).} 
    \centering
    \begin{tabular}{c|cccc|cccc|cccc}
        \toprule
        \multirow{3}{*}{Dataset} 
        & \multicolumn{4}{c|}{DLinear} & \multicolumn{4}{c|}{PatchTST} & \multicolumn{4}{c}{iTransformer} \\
        & \multicolumn{2}{c}{w/o individual} & \multicolumn{2}{c|}{LCD-linear} & \multicolumn{2}{c}{w/o individual} & \multicolumn{2}{c|}{LCD-linear} & \multicolumn{2}{c}{w/o individual} & \multicolumn{2}{c}{LCD-linear} \\
        & MSE & MAE & MSE & MAE & MSE & MAE & MSE & MAE & MSE & MAE & MSE & MAE \\
        \midrule
        Weather & 0.162 & 0.268 & \textbf{0.147} & \textbf{0.252} & 0.163 & 0.270 & \textbf{0.139} & \textbf{0.242} & 0.157 & 0.259 & \textbf{0.139} & \textbf{0.242} \\
        Exchange & 0.083 & 0.214 & \textbf{0.082} & \textbf{0.209} & 0.088 & 0.217 & \textbf{0.087} & \textbf{0.215} & 0.093 & 0.230 & \textbf{0.082} & \textbf{0.209} \\
        ETTh1 & 0.452 & 0.448 & \textbf{0.441} & \textbf{0.445} & 0.450 & 0.451 & \textbf{0.433} & \textbf{0.453} & 0.530 & 0.448 & \textbf{0.441} & \textbf{0.445} \\
        ETTh2 & \textbf{0.183} & \textbf{0.293} & 0.187 & 0.295 & \textbf{0.183} & \textbf{0.294} & 0.194 & 0.298 & \textbf{0.178} & \textbf{0.290} & 0.186 & 0.291 \\
        Traffic-s & 0.587 & 0.446 & \textbf{0.529} & \textbf{0.424} & 0.361 & 0.313 & \textbf{0.345} & \textbf{0.302} & 0.349 & 0.300 & \textbf{0.318} & \textbf{0.288} \\
        ECL-s & 0.342 & 0.392 & \textbf{0.312} & \textbf{0.378} & \textbf{0.261} & \textbf{0.338} & 0.273 & 0.347 & 0.241 & 0.324 & \textbf{0.231} & \textbf{0.316} \\
        \midrule

        & \multicolumn{2}{c}{w/o individual} & \multicolumn{2}{c|}{LCD-as} & \multicolumn{2}{c}{w/o individual} & \multicolumn{2}{c|}{LCD-as} & \multicolumn{2}{c}{w/o individual} & \multicolumn{2}{c}{LCD-as} \\
        \midrule
        Weather & \textbf{0.141} & \textbf{0.237} & 0.145 & 0.248 & 0.150 & 0.253 & \textbf{0.142} & \textbf{0.247} & 0.152 & 0.257 & \textbf{0.147} & \textbf{0.249} \\
        Exchange & 0.081 & 0.208 & \textbf{0.079} & \textbf{0.207} & 0.091 & 0.223 & \textbf{0.081} & \textbf{0.208} & 0.090 & 0.226 & \textbf{0.089} & \textbf{0.225} \\
        ETTh1 & 0.440 & 0.454  & \textbf{0.434} & \textbf{0.451} & 0.433 & 0.460 & \textbf{0.425} & \textbf{0.452} & \textbf{0.443} & 0.467 & 0.444 & \textbf{0.460} \\
        ETTh2 & \textbf{0.182} & 0.294 & 0.184 & \textbf{0.291} & 0.196 & 0.309 & \textbf{0.180} & \textbf{0.290} & 0.194 & 0.307 & \textbf{0.187} & \textbf{0.295} \\
        Traffic-s & 0.615 & 0.474 & \textbf{0.535} & \textbf{0.423} & 0.359 & 0.309 & \textbf{0.344} & \textbf{0.301} & 0.322 & 0.287 & \textbf{0.320} & \textbf{0.285} \\
        ECL-s & 0.365 & 0.412 & \textbf{0.320} & \textbf{0.381} & \textbf{0.260} & \textbf{0.338} & 0.275 & 0.350 & 0.232 & 0.316 & \textbf{0.229} & \textbf{0.314} \\
        \bottomrule
    \end{tabular}
    \label{lcd-wo-indiv}
\end{table*}

\begin{table*}[tbp]\rmfamily
    \caption{Comparison of LCD and its variant using original $\bm{x}$ when predicting $\bm{S}$ ($H = 96$).} 
    \centering
    \begin{tabular}{c|cccc|cccc|cccc}
        \toprule
        \multirow{3}{*}{Dataset} 
        & \multicolumn{4}{c|}{DLinear} & \multicolumn{4}{c|}{PatchTST} & \multicolumn{4}{c}{iTransformer} \\
        & \multicolumn{2}{c}{w/o centering} & \multicolumn{2}{c|}{LCD-linear} & \multicolumn{2}{c}{w/o centering} & \multicolumn{2}{c|}{LCD-linear} & \multicolumn{2}{c}{w/o centering} & \multicolumn{2}{c}{LCD-linear} \\
        & MSE & MAE & MSE & MAE & MSE & MAE & MSE & MAE & MSE & MAE & MSE & MAE \\
        \midrule
        Weather & \textbf{0.141} & \textbf{0.240} & 0.147 & 0.252 & 0.141 & 0.243 & \textbf{0.139} & \textbf{0.242} & 0.140 & 0.244 & \textbf{0.139} & \textbf{0.242} \\
        Exchange & 0.089 & 0.220 & \textbf{0.082} & \textbf{0.209} & 0.095 & 0.220 & \textbf{0.087} & \textbf{0.215} & 0.083 & 0.212 & \textbf{0.082} & \textbf{0.209} \\
        ETTh1 & 0.446 & 0.448 & \textbf{0.441} & \textbf{0.445} & \textbf{0.425} & \textbf{0.445} & 0.433 & 0.453 & 0.442 & 0.446 & \textbf{0.441} & \textbf{0.445} \\
        ETTh2 & 0.208 & 0.313 & \textbf{0.187} & \textbf{0.295} & \textbf{0.191} & \textbf{0.297} & 0.194 & 0.298 & \textbf{0.185} & 0.292 & 0.186 & \textbf{0.291} \\
        Traffic-s & 0.619 & 0.469 & \textbf{0.529} & \textbf{0.424} & 0.411 & 0.363 & \textbf{0.345} & \textbf{0.302} & 0.347 & 0.311 & \textbf{0.318} & \textbf{0.288} \\
        ECL-s & 0.345 & 0.403 & \textbf{0.312} & \textbf{0.378} & 0.303 & 0.375 & \textbf{0.273} & \textbf{0.347} & 0.244 & 0.328 & \textbf{0.231} & \textbf{0.316} \\
        \midrule

        & \multicolumn{2}{c}{w/o centering} & \multicolumn{2}{c|}{LCD-as} & \multicolumn{2}{c}{w/o centering} & \multicolumn{2}{c|}{LCD-as} & \multicolumn{2}{c}{w/o centering} & \multicolumn{2}{c}{LCD-as} \\
        \midrule
        Weather & 0.147 & 0.251 & \textbf{0.145} & \textbf{0.248} & 0.142 & \textbf{0.243} & \textbf{0.142} & 0.247 & \textbf{0.145} & \textbf{0.247} & 0.147 & 0.249 \\
        Exchange & 0.457 & 0.372 & \textbf{0.079} & \textbf{0.207} & 0.297 & 0.334 & \textbf{0.081} & \textbf{0.208} & 3.761 & 0.876 & \textbf{0.089} & \textbf{0.225} \\
        ETTh1 & 0.589 & 0.547  & \textbf{0.434} & \textbf{0.451} & 0.497 & 0.494 & \textbf{0.425} & \textbf{0.452} & 0.481 & 0.482 & \textbf{0.444} & \textbf{0.460} \\
        ETTh2 & 0.209 & 0.313 & \textbf{0.184} & \textbf{0.291} & 0.203 & 0.311 & \textbf{0.180} & \textbf{0.290} & 0.213 & 0.317 & \textbf{0.187} & \textbf{0.295} \\
        Traffic-s & 0.612 & 0.468 & \textbf{0.535} & \textbf{0.423} & 0.397 & 0.309 & \textbf{0.344} & \textbf{0.301} & 0.348 & 0.305 & \textbf{0.320} & \textbf{0.285} \\
        ECL-s & 0.508 & 0.516 & \textbf{0.320} & \textbf{0.381} & 0.314 & 0.378 & \textbf{0.275} & \textbf{0.350} & 0.244 & 0.329 & \textbf{0.229} & \textbf{0.314} \\
        \bottomrule
    \end{tabular}
    \label{lcd-wo-centering}
\end{table*}

\subsection{LCD Performance}
\label{LCDp}

\textbf{Enhancements over SOTA forecasting models.} We evaluated three contemporary state-of-the-art models: the linear model DLinear, the patching model PatchTST, and the Transformer model iTransformer. The results in \ref{lcd-imp} demonstrate that the LCD consistently enhances the performance of all models on all datasets\footnote{Traffic-s (Traffic-small) and ECL-s (Electricity-small) retain only the first 30 dimensions of the Traffic and Electricity datasets, respectively. This is because PatchTST employs a channel-independent approach, which can lead to memory overflow when forecasting high-dimensional time series.} Specifically, LCD-linear reduces the average Mean Squared Error (MSE) of DLinear by $6.29\%$, indicating that even when prediction and normalization models are of the same type, performance can still be improved by addressing inner-instance distribution shifts. In the case of PatchTST and iTransformer, LCD-linear decreases the average prediction error by $6.80\%$ and $8.38\%$, respectively. Furthermore, LCD-as reduces the MSE of three models $6.93\%$, $8.69\%$, and $6.17\%$, thus underscoring the effectiveness of LCD in improving forecast performance. On DLinear and PatchTST, the two LCD methods demonstrate comparable forecasting performance, yet LCD-linear proves superior to LCD-as in Transformer-based models.

\textbf{Comparison with SOTA Normalization Models.} We conducted a comparative analysis of LCD-linear against three normalization models designed to mitigate inter-space shifts: Dish-TS, an instance-level model, SAN, a slice-level model, and Non-stationary Transformer NST. The results in \ref{Table5} demonstrate that LCD-linear significantly outperforms both SAN and Dish-TS. Furthermore, SAN generally exhibits superior performance compared to Dish-TS, indicating that finer-grained normalization strategies yield better outcomes. On average, LCD-linear reduces the state-of-the-art SAN model error by approximately $10\%$. Given that SAN employs a two-stage training schema, while LCD can be seamlessly integrated into any model framework for training purposes, LCD-linear not only delivers high performance but also offers significant convenience. Visualization in \ref{fig-lcd-san-dishts} shows that finer-grained point-level LCD outperforms slice-level SAN and instance-level Dish-TS by addressing inner shifts.    

\textbf{Comparison with variants of LCD.} \ref{lcd-wo-indiv} shows comparison of LCD and w/o individual, which learns shared $\bm{f}$ and $h$ for all features. Using a shared forecasting function implicitly posits that different features share the same time-dependent relationships, which is a strong assumption. The results show that w/o individual improves performance only in a few cases. The preliminary conclusion is that the choice of w/o individual largely depends on the data, and a shared function can only be used when the features of the dataset exhibit high correlation. \ref{lcd-wo-centering} displays the comparison experimental results between when centered inputs $\bm{x} - \mu$ and original inputs are used as inputs to the neural network $\bm{S}$. The results indicate that using centered lookback as input yields better results. 

\subsection{Effectiveness of the Point-Level Paradigm}
\label{effectivenss-tpl}

Both LD and LCD are time-point-level normalization techniques. To assess the effectiveness of the time-point-level mechanism, we compared LD and LCD-linear with their instance-level counterparts: instance-level Learning Distribution and instance-level LCD-linear. Specifically, instance-level LD focuses on fitting the common mean for all different time steps, letting $\bm{A} \in R^D$ and $\bm{P} \in R^D$. Instance-level LCD-linear predicts a shared scaling coefficient $\bm{S} \in R^D$ for all time steps. The experimental results, detailed in \ref{Table7} and \ref{Table8}, indicate that LD performs slightly better than instance-level LD, while LCD-linear surpasses instance-level LCD-linear by approximately $11\%$. This further reinforces the assertion that the advantages of our models stem from effectively addressing inner-instance shifts via the point-level approach.

\section{Conclusion}

In this paper, we introduce a novel category of distribution shift\textemdash inner-instance distribution shift. Based on the characteristics of inner-instance shift, we propose several point-level normalization approaches. To demonstrate the effectiveness of point-level normalization, we respectively compare it with previous methods: we propose LD as a counterpart to normalization methods that use learnable parameters to fit distributions, and LCD as a counterpart to approaches employing neural network architectures to predict statistical metrics. Experiments are conducted using real-world data from multiple domains on different backbone models of various families. The results show that the point-level method outperforms conventional instance-level and slice-level normalization techniques, demonstrating that finer-grained approaches can simultaneously address both inner-instance and instance-level distribution disparities, thereby achieving superior performance. Our proposed normalization methods are lightweight plug-and-play networks, which can be directly applied in practice or integrated as a modular component into future models developed by other researchers. Given the widespread prevalence of inner-instance shifts, we hope that our work would inspire more researchers to investigate inner-instance normalization methods.

\printcredits

\bibliographystyle{cas-model2-names}

\bibliography{main}

\end{document}